%% file: main.tex
\documentclass{article}
\usepackage[numbers]{natbib}
\usepackage{graphicx}
\usepackage{multirow}
\usepackage{array}
\usepackage{tcolorbox}
\usepackage{xcolor}
\usepackage{caption}
\usepackage{longtable}
\newtcolorbox{greybox}{colback=gray!10}
\usepackage{rotating} 
\usepackage{makecell}   
\usepackage{xcolor}
\usepackage{tabularray}
\usepackage{tabularx}
\usepackage{pdflscape}
\usepackage{etoolbox}
\usepackage{float}
\newcounter{magicrownumbers}

\newcommand{\rownumber}{%
  \stepcounter{magicrownumbers}\arabic{magicrownumbers}%
}

\newcommand{\TB}{\textbullet}

\AtBeginEnvironment{longtblr}{\setcounter{magicrownumbers}{0}}

\usepackage{arxiv}
\usepackage[utf8]{inputenc} 
\usepackage[T1]{fontenc}    
\usepackage{url}            
\usepackage{booktabs}       
\usepackage{amsfonts}       
\usepackage{nicefrac}       
\usepackage{microtype}      
\usepackage{graphicx}
\usepackage{doi}
\usepackage{hyperref}

\title{On the Design and Evaluation of Human-centered Explainable AI Systems: A Systematic Review and Taxonomy}

\author{ \href{https://orcid.org/0009-0006-8240-0922}{\includegraphics[scale=0.06]{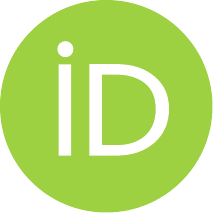}\hspace{1mm}Aline ~Mangold} \\
	Department Speculative Transformation\\
	Dresden University of Technology\\
	Dresden, 01062 \\
	\texttt{aline.mangold@tu-dresden.de} \\
	\And
	\href{https://orcid.org/0009-0009-0132-4473}{\includegraphics[scale=0.06]{orcid.pdf}\hspace{1mm} Juliane ~Zietz} \\
	AG Translational Computational Psychiatry\\
	Charité Berlin\\
	Berlin, 10117 \\
	\texttt{juliane.zietz@charite.de} \\
	\AND
    \href{https://orcid.org/0009-0006-0892-5712}{\includegraphics[scale=0.06]{orcid.pdf}\hspace{1mm} Susanne ~Weinhold} \\
	Chair of Industrial Design Engineering\\
    Dresden University of Technology\\
	Dresden, 01062 \\
	\texttt{susanne.weinhold@tu-dresden.de} \\
	\And
	\href{https://orcid.org/0000-0002-6673-9591}{\includegraphics[scale=0.06]{orcid.pdf}\hspace{1mm} Sebastian ~Pannasch} \\
	Chair of Engineering Psychology and Applied Cognitive Research \\
    Dresden University of Technology\\
	Dresden, 01062 \\
	\texttt{sebastian.pannasch@tu-dresden.de} \\
}

\renewcommand{\shorttitle}{On the Design and Evaluation of Human-centered Explainable AI Systems}

\hypersetup{
pdftitle={Review Paper},
pdfsubject={cs},
pdfauthor={Aline ~Mangold, Juliane ~Zietz, Susanne ~Weinhold, Sebastian ~Pannasch},
pdfkeywords={Explainable artificial intelligence (XAI), human-computer interaction (HCI), machine learning, explanation, transparency, human-centered},
}

\begin{document}
\maketitle

\begin{abstract}
\noindent As AI becomes more common in everyday living, there is an increasing demand for intelligent systems that are both performant and understandable. Explainable AI (XAI) systems aim to provide comprehensible explanations of decisions and predictions. At present, however, evaluation processes are rather technical and not sufficiently focused on the needs of human users. Consequently, evaluation studies involving humans can serve as a valuable guide for conducting user studies.
This paper presents a comprehensive review of 65 user studies evaluating XAI systems across different domains and application contexts. As a guideline for XAI developers, we provide a holistic overview of the properties of XAI systems and evaluation metrics focused on users (human-centered). We propose goals for the human-centered design (design goals) of XAI systems for different user groups (AI novices and data experts).
The first part of our results includes the analysis of XAI system characteristics. Importantly, we distinguish between the core system and the XAI explanation, which together form the whole system. Further results include the distinction of evaluation metrics into affection, cognition, usability, interpretability, and explanation metrics. Furthermore, the users, along with their specific characteristics and behavior, can be assessed. For AI novices, the relevant extended design goals include responsibility, acceptance and user experience. For data experts, the focus is performance-oriented, including human-AI collaboration and system task performance. Future literature trends include the design of context-aware and inclusive systems, adapting to users and the situational context. Several limitations in the reviewed literature were identified: a lack of validation, standardization, and documentation; evaluation is focused on particular metrics, rather than holistic; evaluation of explanatory and transparency components of the system is often neglected in user studies. These research gaps could be addressed by integrating existing XAI evaluation frameworks and developing validated questionnaires.
\end{abstract}

\noindent\textbf{Keywords: }\textit{{Explainable artificial intelligence (XAI), human-computer interaction (HCI), machine learning, explanation, transparency, human-centered}}

\maketitle

\input{sections/introduction}
\input{sections/background}
\input{sections/methodology}
\input{sections/results}
\input{sections/discussion}
\input{sections/conclusion}

\section{Conflict of Interest Statement}
The authors declare that they have no known competing financial interests or personal relationships that could have appeared to
influence the work reported in this paper.
\section{Declaration of AI Use}
AI–based tools were used in the preparation of this manuscript solely for language enhancement purposes. Specifically, ChatGPT, DeepL, and Grammarly were employed to assist with grammar correction, phrasing, and overall clarity of the text. These tools were not used to generate original research content, data, analyses, or interpretations. All substantive intellectual contributions, including the study design, analysis, and conclusions, remain the responsibility of the authors.

\clearpage
\bibliography{PhD_Aline.bib}         
\newpage
\input{sections/appendices}

\end{document}

%% file: sections/introduction.tex
\section{Introduction}
Over the past decade, Artificial Intelligence (AI) has rapidly spread across various sectors \cite{gozalo-brizuelaSurveyGenerativeAI2023}, including, healthcare \cite{jiangArtificialIntelligenceHealthcare2017}, marketing \cite{nairApplicationAITechnology2021}, and finance \cite{caoAIFinanceChallenges2022}. Therefore, it is becoming an integral part of high-stake decision-making tasks such as medical diagnosis, personal data analysis, and credit scoring. These applications have made AI an indispensable tool, but have also raised concerns about its influence and the potential consequences of its misuse \cite{rudinWhyBlackBox2022}. For example, when confronted with inaccurate AI recommendations, healthcare professionals may make incorrect diagnostic decisions \cite{jussupowAugmentingMedicalDiagnosis2021}, which can be risky for patients.
\par
Bender \cite{benderDangersStochasticParrots2021} emphasized AI's risk for further societal impacts, such as bias or misinformation. To mitigate these effects, there is a growing demand for AI systems to be accountable, controllable, and explainable \cite{fjeldPrincipledArtificialIntelligence2020}. Explainability aims to help end users understand how systems arrive at their predictions or recommendations to promote safe use in critical applications \cite{waddenDefiningUndefinableBlack2022}. Furthermore, explainable algorithms can reveal unfairness in decision-making processes related to socio-demographic characteristics \cite{zhaoFairnessExplainabilityBridging2022} and misinformation in social media posts \cite{gongIntegratingSocialExplanations2024}.
Despite the demand for explainability, "Black-box models" are currently the dominant approach in AI \cite{kamathExplainableArtificialIntelligence2021}. Black-box models, such as deep learning models, lack an interpretable structure and explanation that describes how their output was produced.
Although some models like decision trees are more easily interpretable by users and can be designed transparently (white-box models) \cite{aliExplainableArtificialIntelligence2023}, black-box models often provide higher predictive accuracy. This leads to a trade-off between accuracy and transparency \cite{kamathExplainableArtificialIntelligence2021}.
\par
Since black-box models are commonly used and provide great potential in output quality it is important to gain a deeper understanding of their inner workings. This need was addressed by the Defense Advanced Research Projects Agency (DARPA) \cite{gunningDARPAsExplainableArtificial2019}, which introduced the concept of explainable AI (XAI). The DARPA XAI program aimed to advance explainable machine learning (ML) by exploring three primary methods. These are: adapting deep learning to produce interpretable features; creating structured causal models; and using model induction to derive explainable versions of black-box models.
\par
Building on this program, various XAI algorithms have emerged \cite{chenOverviewXAIAlgorithms2023}. However, even though XAI methods like Local Interpretable Model-Agnostic Explanations (LIME) \cite{ribeiroWhyShouldTrust2016} have gained traction, they remain predominantly suited for AI developers rather than end-users. Introduced in 2016, LIME offers a way to interpret predictions from any machine learning model, by creating locally faithful explanations for individual predictions. This approach is particularly valuable for developers in debugging or validating models by analyzing how predictions shift with small input changes \cite{visaniStatisticalStabilityIndices2020}. However, outputs produced by LIME contain various technical terms and mathematical expressions that could impede end-user understanding.
\par
Another issue is that XAI methods are not only primarily targeted toward developers but are often only evaluated from a technical point of view. This means that they do not incorporate users of any kind \cite{nautaAnecdotalEvidenceQuantitative2023}. For instance, when assessing XAI from a technical standpoint, factors such as correctness (the degree to which the explanation aligns with the predictive model) and completeness (the extent to which the explanation clarifies the predictive model) can be considered. From a user perspective, on the other hand, metrics such as explanation satisfaction and accurate user mental models could be useful indicators of explanation quality \cite{hoffmanMetricsExplainableAI2019a}.
As noted in the literature \cite{sentHumancenteredEvaluationExplainable2024}, there is indeed a distinction between XAI evaluations that are \textbf{computer-centered} \cite{lopesXAISystemsEvaluation2022} and those that are \textbf{human-centered}. Computer-centered evaluations do not incorporate users. In contrast, human-centered evaluation methods are assessed with users in the context of use, which helps to incorporate their perspective.
\par
\begin{figure}[H]
		\caption{User Groups of XAI adapted from Mohseni \cite{mohseniMultidisciplinarySurveyFramework2021}.}
		\includegraphics[width=14cm]{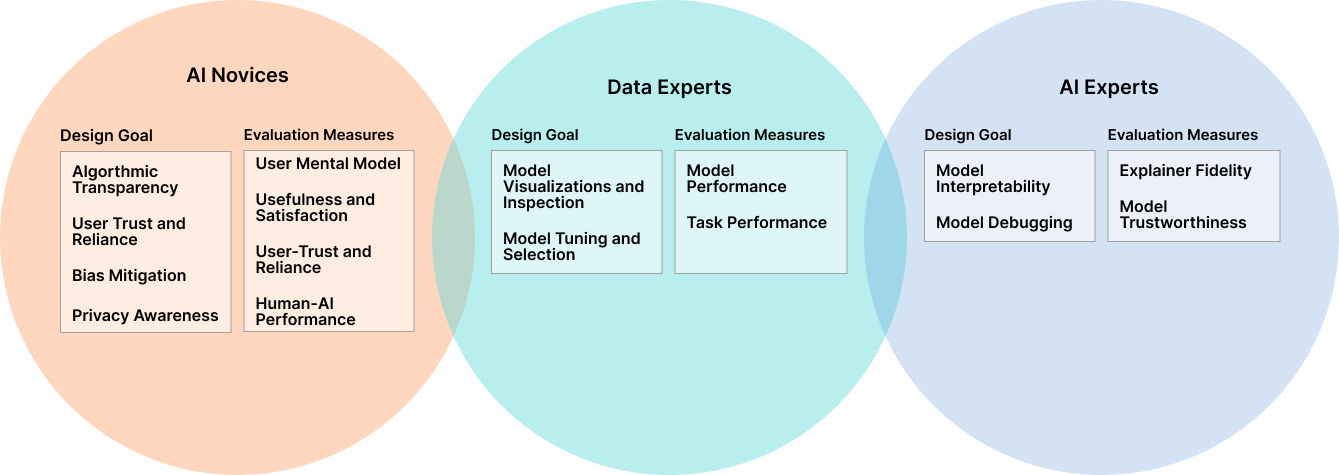}
		\label{fig: user_groups_old}
	\end{figure}

The evaluation of systems is an essential part of various development frameworks. For instance, the design science research process \cite{hevnerThreeCycleView2007} describes the development of systems as an iterative procedure with several design cycles. As evaluation studies are placed at the end of each design cycle, the development of design artifacts and processes occurs first. To align this step with the needs of XAI system users, human-centered design goals must be defined before development. Mohseni \cite{mohseniMultidisciplinarySurveyFramework2021} identified three different user groups for XAI: AI novices, data experts, AI experts. AI experts design ML algorithms and interpretation techniques for XAI systems; data experts use ML for analysis, making decisions, or conducting research, and AI novices are end-users who use AI products in their daily lives but have little to no experience with ML systems. Besides this distinction in XAI user groups, Mohseni \cite{mohseniMultidisciplinarySurveyFramework2021} provided design goals for each group, see figure ~\ref{fig: user_groups_old}.  While AI experts and data experts use XAI for model debugging and inspection, AI novices may prioritize ethical and affective considerations such as transparency, trust, and reliance. Thus, when designing XAI systems for different user groups, it is important to consider the specific design goals for each group. For instance,  user trust can be improved by dialogically presented explanations \cite{millerExplanationArtificialIntelligence2019}. In contrast, when developing systems that should assist users in model debugging, XAI explanations should include various summary statistics \cite{huangAnalysisDevelopmentXAI2022}. Nevertheless, rather than solely focusing on the explanation in XAI design, the system needs to be considered in its entirety. For instance, Shneiderman \cite{shneidermanDesignLessonsAIs2020} proposed two approaches to AI design: emulation (mimicking human abilities), and the creation of useful applications that solve real-world problems. An XAI explanation can help the user understand the system and use it more effectively, but it cannot solve user tasks. Thus, designing for real-world problems can only be achieved by an XAI system as a whole, not solely by an XAI explanation.
In this paper, we contribute to the adoption of human-centered design and evaluation practices by providing relevant design goals for distinct user groups (AI novices and data experts) and possible evaluation metrics for XAI systems. AI experts, which present the third user group of XAI were not found as a target group in the reviewed literature and are therefore no subject of this paper. Our approach extends prior evaluation and design frameworks (e.g. \cite{mohseniMultidisciplinarySurveyFramework2021}) by adding design goals and a holistic overview of human-centered ai metrics. Furthermore, we provide guidance on selecting appropriate metrics, validated questionnaires, and best practices for conducting user studies.
To this end, we review 65 user studies of XAI systems across various domains and application contexts. Our review makes three key contributions to the field. 
\begin{itemize}
    \item First, it provides a thorough overview of human-centered evaluation measures currently employed in XAI research.
    \item  Second, it pairs these evaluation measures with relevant design goals for two user groups (AI novices and data experts) of XAI systems.
    \item Third, it extends existing XAI evaluation taxonomies derived from previous reviews.
    \item  Lastly, it identifies key research gaps and areas for improvement within XAI research and provides recommendations and guidelines to improve it.
\end{itemize}
The structure of this paper is as follows: Section ~\ref{sec:background} provides a detailed explanation of XAI and related concepts, followed by an overview of current XAI evaluation practices and related literature. Section ~\ref{sec: review method} outlines our methodology. In section ~\ref{sec:descriptive results} , we introduce our descriptive results, including paper publication dates, application domains, and participants. In Section ~\ref{sec: synthesis}, we present the synthesis of the concepts identified in the literature: Section ~\ref{sub-sec: Taxonomy of XAI System Properties and Evaluation Metrics} introduces our categorization of XAI systems and evaluation metrics, and Section ~\ref{sub-sec: Design Goals} includes our design goals adapted to AI novices and data experts.
Section ~\ref{sec:discussion} summarizes our findings and discusses the limitations and possibilities for extending our taxonomy and future research. In Section 7, we present a conclusion to our work.

%% file: sections/background.tex
\section{Background}
\label{sec:background}
Despite its potential to explain black-box models, the research community does not agree on what exactly constitutes XAI \cite{palacioXAIHandbookUnified2021}. Furthermore, many associated terms around the concept of XAI also lack agreement. Therefore, to ensure an effective review and discussion of XAI evaluation practices, it is crucial to define XAI and its related terms that align with our specific context.
\subsection{XAI}
Machine learning explanations are increasingly essential in critical applications such as healthcare or finance. For instance, in the medical field, XAI helps clinicians understand AI predictions in diagnoses and treatments \cite{sheuSurveyMedicalExplainable2022}. In finance, XAI provides transparency in complex predictive models for tasks like credit scoring \cite{delangeExplainableAICredit2022}. XAI systems aim to clarify the reasoning behind model decisions, helping users to understand data processing, detect bias, and identify system issues \cite{alikhademiCanExplainableAI2021}. However, despite its increased usage and associated benefits, a consensus on what constitutes XAI is still pending, and various definitions exist. For example, Gunning \cite{gunningDARPAsExplainableArtificial2019} predicted that \textit{“XAI will create a suite of machine learning techniques that enables human users to understand, appropriately trust, and effectively manage the emerging generation of artificially intelligent partners”}, whereas Barredo Arrieta \cite{barredoarrietaExplainableArtificialIntelligence2020} defines XAI as following:
\textit{"Given an audience, an explainable Artificial Intelligence is one that produces details or reasons to make its functioning clear or easy to understand."}.
The first definition conceptualizes XAI as a set of machine learning techniques. In contrast, the second definition removes this technical focus and emphasizes the adaptability of an AI system's explainability according to the needs and capabilities of its audience. However, as we deal with human users, in our context, XAI does refer to:
\begin{greybox}
    \textit{AI systems that provide user explanations of model predictions or overall model functioning understandably}.
\end{greybox}
This description differs from the aforementioned definitions concerning two aspects: (i) The generation of XAI explanations is not necessarily dependent on machine learning algorithms, and (ii) the intended audience for XAI explanations are human users. Since \textit{explanations} are an essential part of our understanding of XAI, we want to elaborate on this concept. The term "explanation" can be used in interaction contexts between two people. For example, Schmid \cite{schmidMutualExplanationsCooperative2020} outline that \textit{“in human–human interaction, explanations have the function to make something clear by giving a detailed description, a reason, or justification"}, while  Palacio \cite{palacioXAIHandbookUnified2021} suggest that \textit{"an explanation is the process of describing one or more facts, such that it facilitates the understanding of aspects related to said facts (by a human consumer)."} Other definitions focus on the explanation of the ML model by saying that an explanation is a \textit{“presentation of (aspects of) the reasoning, functioning and/or behavior of a machine learning model in human-understandable terms.”} \cite{nautaAnecdotalEvidenceQuantitative2023}. The latter definition is interesting since only the consumer is human, but the transmitter of the explanations is not further defined; furthermore,  "facts" are specified as "reasoning, functioning and behavior". For our context, we provide the following definition for explanations:
\begin{greybox}
\textit{An explanation is provided by an XAI system (sender) and is the product of a process that describes the internal functioning of the model in such a way that it is understandable by the user (receiver) and makes it easier for them to comprehend the model}.
\end{greybox}
\subsection{XAI Terminology}
\label{sub-sec:XAI Terminology}
In addition to the overarching concept of XAI, several related sub-terms have emerged that also lack consensus. The initial step of this review was a scan of prominent XAI literature reviews to gain an overview of the current research landscape and identify relevant research gaps \cite{brendelWhatLiteratureReview2020}. In this regard, we extracted significant XAI-related terms that could form part of our search query in subsequent steps of the systematic literature review. These terms represent the target properties of XAI: \textbf{explainability}, \textbf{interpretability}, \textbf{transparency}, and \textbf{understandability (equivalent with intelligibility)}. For the sake of clarity, we have provided definitions from the literature, but it should be noted that these terms will be redefined later based on the literature we have reviewed.
The target properties of XAI, as outlined in table ~\ref{tab:target_properties_xai}, are introductory descriptions that represent partly overlapping constructs and vary depending on the context. For this reason, these terms are often used interchangeably by researchers \cite{adadiPeekingBlackboxSurvey2018}.
\input{Tables/XAI_Target_Properties}
\newpage
\subsection{Current XAI Evaluation Practices}
To assess the target properties of XAI, as outlined in section ~\ref{sub-sec:XAI Terminology}, it is necessary to conduct evaluation studies. However, at present, the assessment of XAI systems is predominantly technical \cite{nautaAnecdotalEvidenceQuantitative2023}. The authors found that only one out of five papers examined included a user evaluation of the XAI system. Moreover, recent surveys \cite{adadiPeekingBlackboxSurvey2018, anjomshoaeExplainableAgentsRobots2019, doshi-velezRigorousScienceInterpretable2017} have identified shortcomings in current XAI user evaluations. Anjomshoae \cite{anjomshoaeExplainableAgentsRobots2019} reviewed  62 studies and found that while 97\% of the studies recognized the need for user explanations, 41\% lacked user evaluation, and many failed to report and discuss their findings comprehensively. Adadi \cite{adadiPeekingBlackboxSurvey2018} analyzed 381 papers and found that only 5\% explicitly focused on XAI evaluation. Furthermore, Doshi-Velez \cite{doshi-velezRigorousScienceInterpretable2017} identified a lack of standardization regarding the evaluation of interpretability across specific user contexts and tasks. Thus, they emphasized that explanations need to meet user needs across different domains. Their proposed framework also recommends measuring interpretability through behavioral outcomes rather than subjective opinions alone. Subsequently, researchers need to develop quantitative behavioral benchmarks to assess the effectiveness of explanations objectively. Refer to appendix \ref{app:prior_work} for an overview of existing taxonomies.
\par
Building on this early work, a series of literature reviews has attempted to consolidate the fragmented XAI evaluation landscape, though none has fully addressed it from an integrated, human-centered perspective. Mohseni \cite{mohseniMultidisciplinarySurveyFramework2021} proposed concept matrices and design goals from both a computer- and human-centered perspective, forming a key foundation for our review, but they did not report an explicit or systematic review method. Vilone \cite{viloneNotionsExplainabilityEvaluation2021} and Zhou \cite{zhouEvaluatingQualityMachine2021} similarly produced tables and taxonomies of evaluation approaches from a predominantly computer-centered viewpoint, again without a stated review methodology. \cite{lopesXAISystemsEvaluation2022} combined computer- and human-centered evaluation criteria into tabular form, while Nauta \cite{nautaAnecdotalEvidenceQuantitative2023} introduced the Co-12 explanation properties, focusing mainly on computer-centered metrics. More recently, Naveed \cite{naveedOverviewEmpiricalEvaluation2024} conducted an unsystematic scoping review to derive concept matrices and evaluation guidelines focused solely on metrics, and Sent \cite{sentHumancenteredEvaluationExplainable2024} applied a systematic PRISMA-based review to develop a taxonomy centered on the meaningfulness of explanations to users. Across these efforts, review methods are rarely made explicit, and each review narrows its focus to a specific facet of evaluation (e.g., metrics, meaningfulness, or explanation properties) rather than integrating them. For a detailed overview, please refer to appendix \ref{app:prior_work}.
Our review addresses these gaps by asking how existing XAI evaluation metrics can be integrated to enable a human-centered evaluation of XAI systems, and which design goals can be derived from the existing literature. Unlike prior work, it applies a composite, structured literature review method and produces a combined set of artefacts: concept matrices, design goals, evaluation and design guidelines, and a taxonomy. 
\subsection{Related Reviews and Guidelines}
In response to the shortcomings of human-centered evaluation of XAI systems, several reviews have been conducted, resulting in different taxonomies. Kim \cite{sentHumancenteredEvaluationExplainable2024} provided a comprehensive overview of these taxonomies on a high level. We extended this overview by adding Naveed's \cite{naveedOverviewEmpiricalEvaluation2024} distinction of test scenarios and Kim's \cite{sentHumancenteredEvaluationExplainable2024} XAI evaluation levels in table ~\ref{tab:taxonomies_in_XAI_evaluation}. 
In the current XAI evaluation literature, \textbf{human-centered} approaches with users and \textbf{computer-centered} approaches without users distinguish XAI evaluation on a high level. This paper will focus on \textbf{human-centered} approaches since we want to investigate evaluation approaches with human users. The overall distinction of the two approaches answers the question of \textit{who} evaluates the system and in \textit{what context of use} it is evaluated. Kim \cite{sentHumancenteredEvaluationExplainable2024} further addressed the question of \textit{Which aspect of the system is evaluated} by introducing the distinction between the evaluation of the \textit{quality of explanations} and the \textit{contribution of the explanation to the user experience}. The first includes aspects directly related to the explanation, such as the understanding of the explanation or the ease of understanding the explanation. The second one focuses on the remaining parts of the system and includes aspects like satisfaction with the system or the usefulness of the system.
\par 
Doshi-Velez \cite{doshi-velezRigorousScienceInterpretable2017} distinguishes evaluation approaches based on \textit{Who} evaluates the system and \textit{Which tasks} are used in the evaluation. Functionality-grounded evaluations use algorithmic metrics \cite{mohseniQuantitativeEvaluationMachine2021} and formal definitions to assess explanation quality objectively without human participants. \textbf{Application-grounded} evaluations involve expert users in real applications to measure aspects like explanation quality \cite{nautaAnecdotalEvidenceQuantitative2023}. \textbf{Human-grounded} evaluations engage less experienced users to evaluate constructs like understandability, trust, and usability in simple tasks \cite{nautaAnecdotalEvidenceQuantitative2023}. Our reviewed papers both address \textbf{application-grounded} evaluations with domain experts and \textbf{human-grounded} evaluations with lay users. The user groups identified by Mohseni \cite{mohseniMultidisciplinarySurveyFramework2021} can be distinguished by their 
AI usage contexts: no (or very little) expertise and usage of ML-based systems in everyday life (AI Novices), usage of ML for analysis, research, and decision-making (Data Experts), and design of ML algorithms and interpretability techniques (AI Experts). Due to their different tasks, objectives, and characteristics, these user groups require different design goals. According to the taxonomy, AI and data experts have technical tasks, such as model debugging, model tuning, and selection. AI novices need value-driven qualities, such as algorithmic transparency and privacy awareness. Further, specific user behaviors should be encouraged by the XAI design, such as user reliance. Therefore, we define design goals in the following manner:
\begin{greybox}
    \textit{A design goal is a high-level goal that defines what a design is intended to achieve, including the qualities it should embody, the user behaviors it aims to encourage or discourage, and the tasks it should enable users to accomplish}.
\end{greybox}
\par
Besides addressing diverse user groups, XAI evaluation can take place in different scenarios. For instance, \textbf{real-world scenarios} \cite{naveedOverviewEmpiricalEvaluation2024} involve real-world cases in which AI decisions affect individuals or entail a risk of impact on the lives of individuals, groups, or society. These are domains where the stakes of AI decisions are high and require rigorous and reliable explanation systems. \textbf{Illustrative scenarios}, on the other hand, include domains or evaluation scenarios where AI decisions have little impact or where researchers envision simple scenarios to illustrate an approach and the explanations it produces.
\par
Zhou \cite{zhouEvaluatingQualityMachine2021} categorizes evaluation metrics into \textbf{subjective} (e.g., trust, satisfaction) and \textbf{objective} (e.g., task accuracy, gaze fixation) types, focusing on user perception versus measurable indicators. Similarly, Vilone \cite{viloneNotionsExplainabilityEvaluation2021} and Nauta \cite{nautaAnecdotalEvidenceQuantitative2023} distinguish between \textbf{qualitative} and \textbf{quantitative} metrics. 
\par
In table ~\ref{tab:distinction_literature_review} we provide an overview of related XAI evaluation reviews and distinguish them from our work. Most papers differ from ours in that they use their own review method, which in some cases is transparently disclosed, stating the searched databases and search query (e.g., \cite{nautaAnecdotalEvidenceQuantitative2023}). However, in some cases, we could not find adequate information regarding the review process (e.g. \cite{lopesXAISystemsEvaluation2022}). We use the composite literature review method, an approach introduced by Brendel \cite{brendelWhatLiteratureReview2020}, while transparently disclosing all conducted steps. Furthermore, some reviews focused on different perspectives, such as computer-centered and human-centered evaluation practices (e.g. \cite{nautaAnecdotalEvidenceQuantitative2023}). While we solely focus on human-centered practices, we are delving deep into the used metrics in this area. Furthermore, we incorporate a design perspective, building on the work of Mohseni \cite{mohseniMultidisciplinarySurveyFramework2021}. Other reviews focus on evaluation aspects that differ from ours. For instance, Kim \cite{sentHumancenteredEvaluationExplainable2024}investigated the meaningfulness of explanations, while we focus on aspects concerning all components of an XAI system. 

%% file: Tables/XAI_Target_Properties.tex
\begin{table}[h!]
\caption{Target Properties of XAI}
\label{tab:target_properties_xai}
\centering
\renewcommand{\arraystretch}{1.5} 
\begin{tabular}{>{\raggedright}p{4cm}>{\raggedright\arraybackslash}p{4cm}>{\raggedright\arraybackslash}p{4cm}}
\hline
\textbf{Term} & \textbf{Description} & \textbf{Example for Technical Implementation} \\
\hline
\textbf{Explainability} & The ability to explain the underlying model and its reasoning with accurate and user-comprehensible explanations \cite{mohseniMultidisciplinarySurveyFramework2021}. & Local explainers like LIME or SHAP \cite{kamathExplainableArtificialIntelligence2021}
\\
\hline
\textbf{Interpretability} & Interpretability indicates the degree that an AI model becomes clear to humans in a passive way \cite{barredoarrietaExplainableArtificialIntelligence2020}. & Decision trees, linear or logistic regression \cite{kamathExplainableArtificialIntelligence2021}
\\
\hline
\textbf{Transparency} & A level to which a system provides information about its internal workings or structure \cite{tomsettInterpretableWhomRolebased2018}. 
& Model cards, open datasets \cite{mitchellModelCardsModel2019}
\\
\hline
\textbf{Understandability (Intelligibility)} & Characteristic of a model to make a human understand its function – how the model works – without any need for explaining its internal structure or the algorithmic means by which the model processes data internally \cite{barredoarrietaExplainableArtificialIntelligence2020}. & Natural language explanations \cite{zhaoExplainabilityLargeLanguage2024}
\\
\hline
\end{tabular}
\begin{minipage}{\dimexpr 12cm + 6\tabcolsep\relax}
\small\textit{Note:} Due to the varying definitions and interpretations across different literature and application domains, this table is provided as a quick overview and does not claim completeness.
\end{minipage}
\end{table}

%% file: sections/methodology.tex
\section{Review Method}
\label{sec: review method}
This literature study addresses the following research questions:
\begin{quote}\emph{RQ 1: How can existing XAI evaluation metrics be integrated to enable a human-centered evaluation of XAI systems?}\end{quote}
\begin{quote}\emph{RQ 2: Which design goals can be derived from existing XAI evaluation literature?}\end{quote}
To answer these questions, we only considered studies that evaluate XAI systems with human users and extracted relevant aspects of the used evaluation methodology. Subsequently, we will describe the method that was used to identify, select, and analyze the relevant papers.
\subsection{Composite Literature Review Method}
In the current literature, several methods for conducting literature reviews have been proposed \cite{websterAnalyzingPrepareFuture2002}. This inconsistency in conducted literature reviews could lead to a lack of comparability and discrepancy in findings. To address these challenges, Brendel \cite{brendelWhatLiteratureReview2020} provided a synthesis of the most common practices employed in literature reviews. The following subsections will address each step of the methodology, introduced by Brendel \cite{brendelWhatLiteratureReview2020}. This includes \textbf{preparation}, \textbf{scope}, \textbf{search}, and \textbf{analysis}. The results section will address the \textbf{synthesis} step, while the discussion section will refer to the \textbf{discussion} step.
\subsubsection{Preparation}
To conduct a literature review, the first steps are to define the review goal and conceptualize the research field and topics. As outlined in section ~\ref{sec:background}, the initial phase of this literature review entailed a comprehensive examination of existing literature reviews on XAI e.g.\cite{mohseniMultidisciplinarySurveyFramework2021, nautaAnecdotalEvidenceQuantitative2023, barredoarrietaExplainableArtificialIntelligence2020}.
We used these reviews to extract target properties of XAI systems which were used in a first search query.
\subsubsection{Scope}
The next critical step in our process was to narrow the scope of our research. A broad scope could yield results that are not directly relevant to the research question, while a specific scope could exclude important research papers. We narrowed our scope by defining the following \textbf{inclusion criteria}:
\begin{itemize}
    \item evaluation of an \textbf{XAI system}
    \item \textbf{human-centric} evaluation metrics
    \item \textbf{human users} as test subjects
\end{itemize}
Based on our previous findings and inclusion criteria we developed an initial search query.
The databases used for this literature review were:
\begin{itemize}
    \item Institute of Electrical and Electronics Engineers (IEEE Explore)
    \item Association for Computing Machinery (ACM)
\end{itemize}
This focus was chosen because both databases are closely linked to the core areas of computer science and engineering that define XAI research and provide peer-reviewed documents. Compared with larger databases such as Scopus, these databases already concentrate on computer‑science and user‑study research. Consequently, the filtering process is less resource‑intensive because there is far less unrelated material and fewer duplicate records.
We decided to include papers from \textbf{2016-2023}. We chose this starting date because 2016 marks the introduction of the LIME algorithm \cite{ribeiroWhyShouldTrust2016}, a significant milestone in XAI research. The literature search was conducted in January 2024. In former iterations, results were either too broad (providing a high quantity with inadequate literature in pre-screening) or too narrow (yielding very little results). Thus, we pre-tested and adapted our query on the aforementioned databases iteratively until it yielded satisfactory results (an adequate amount of papers matching our inclusion criteria). The final search query was the following: 
\begin{greybox}{
    \textit{("interpretability" OR "explainability" OR "transparency" OR "understandability") AND \newline
    ("AI" OR "artificial intelligence" OR "machine learning" OR "deep learning") AND \newline
    ("XAI" OR "explainable AI" OR "interpretable AI") AND
    \newline
    (test* OR "evaluation") AND \newline
    ("user" OR "non-expert user" OR "domain expert" OR "lay user") AND \newline
    ("usability testing" OR "user experience" OR "effectiveness assessment" OR "user feedback") AND \newline
    ("cognitive load" OR "trust" OR "engagement") AND \newline
    ("qualitative" OR "quantitative").}}
\end{greybox}
\subsubsection{Search}
\label{sub-sub-sec: methodology search}
Using the final query, we conducted a full-text search on ACM (314 results) and IEEE (138 results). As  We filtered the gathered literature in a two-step process:
\begin{enumerate}
    \item Read abstracts and filter based on inclusion criteria.
    \item Read full-texts and define exclusion criteria.
\end{enumerate}
After the first filtering step, we ended up with 116 papers. In the next step, papers were excluded according to the following criteria:
\begin{itemize}
    \item Wrong publication type (e.g., review paper, literature survey, guideline instead of empirical user study; manually excluded)
    \item No XAI system (e.g., AI system without explanatory components)
    \item Broad focus (e.g., Automation, Robotics)
    \item Lack of traceability (e.g,. no reporting of evaluation metrics)
\end{itemize}
After this second step, 65 papers were identified as suitable for review (no duplicates). As the number and quality of papers were satisfactory, no backward search (citation backtracking) was conducted. Brendel \cite{brendelWhatLiteratureReview2020} highlight, that the literature corpus is big enough, if the researcher can make significant and coherent statements about the reviewed field. 
\subsubsection{Analysis}
\begin{figure}[H]
		\caption{Development of the Concept Matrix}
		\includegraphics[width=9cm]{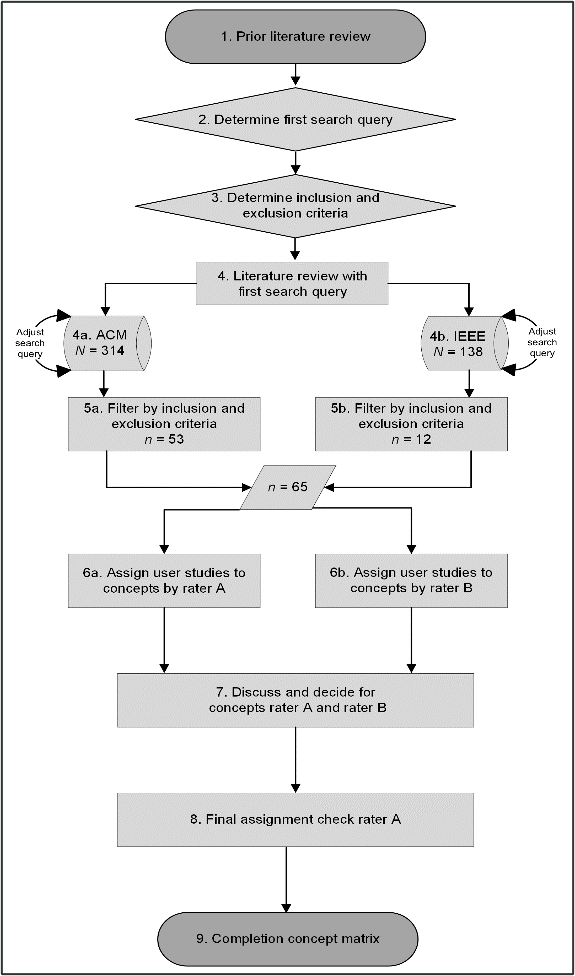}
		\label{fig: concept_matrix_development}
	\end{figure}

Once the relevant papers for the review had been identified, the next step was to extract the key concepts from the selected literature. We applied an inductive approach, which is data-driven \cite{charmazGroundedTheory2007}, to identify relevant concepts. We chose a concept matrix to provide a visual representation of the coded papers. A concept matrix is a structured tool to organize and analyze information by categorizing it into key concepts. The rows represent the items being analyzed (research papers). Columns represent the criteria used to categorize and compare the rows (e.g., a certain evaluation metric). Cells contain the evaluations related to the intersection of a row and a column (presence of a certain characteristic). Two raters independently read and rated the papers to develop an overall concept scheme and assign the concepts to each paper. To guarantee a uniform and impartial rating procedure, definitions for each concept were established. Furthermore, the ratings were conducted following the following coding guidelines:
\begin{itemize}
    \item Only concepts that occur five or more times are included. That is to ensure the relevance of the included concept in the literature. Please find excluded concepts in appendix ~\ref{app:excluded} (occurred less than 5 times).
    \item Evaluation sections with end-users are considered for concept assignment. 
    \item Only the final evaluation section is considered (rather than earlier evaluations of prototype drafts).
    \item If several concepts apply to the same sub-concept, all are assigned. 
    \item No concept assignment without sufficient descriptions.
    \item Name concepts according to our definition.
\end{itemize}
The concept scheme was modified for the first 20\% of the papers (equals the first 13 papers) throughout the process based on discussions between the raters. The final assignment of concepts to papers was conducted through a joint discussion between the raters regarding their initially assigned concepts. Interrater reliability for the first 13 papers was not calculated because the assessments were not completely independent due to the discussions that took place in the meantime. However, the discussions were essential to ensure that the concept and schema modification were formulated in a meaningful way. The remaining 80\% of the papers were rated using the fixed coding scheme. Since the categories were not mutually exclusive (e.g. multiple metrics) no established reliability rating could be calculated. For instance, Cohen's Kappa is a popular reliability metric but its calculation requires mutually exclusive coding schemes \cite{gisevInterraterAgreementInterrater2013}. However, in our case, mutually exclusive codes would be implausible in terms of content. For instance, regarding the overall system, multiple XAI methods could be applied. Considering system evaluation, several metrics could be measured at once.
The final assignment check was conducted by one rater. This was done to verify that the codes recorded in the rater documents matched the concept matrices and that no transfer errors had occurred. Thus, the assignment check did not include any new interpretations or reassignments. Please find an illustration of the concept matrix development process in figure ~\ref{fig: concept_matrix_development}. The analysis yielded two distinct concept matrices. The first matrix characterizes the XAI system itself, comprising 16 concepts that describe its core components and the explanations it generates. The second matrix captures the metrics used in the user study; it contains 19 concepts that address the same core system and explanations while also incorporating concepts that describe the users who interact with the system. Please find the corresponding concept matrices in appendix ~\ref{app:matrix}.
\par
To derive specific design goals for AI novices and data experts, each paper was reviewed again by two raters. The user group was considered AI novice, if there was either no remark about a technical background, they were described as lay-users or users were considered experts from other fields (e.g. medical staff or designers). The user group was considered data expert, if there was information about users' background in this regard (e.g. profession as data scientist). In particular, the sections addressing the research questions of the evaluation study were analyzed. Design goals were formulated by synthesizing reoccurring key words such as calibrated trust, acceptance, user experience, performance, or human-AI collaboration to formulate overarching design goals. Design goals were formulated in a way that left minimal possible overlap with other design goals. For each paper, we documented the occurrence of design goals. Then, we recorded the design goals for each user group depending on which user group was addressed in the paper. Similarly to the development of the concept matrices, no mutual exclusive coding was possible, as one paper could cover multiple design goals. Lastly, we assigned the design goals to the user group based on their relative occurrence. That means, that a design goal was assigned to a user group if it had a higher relative frequency in this group than in the other group. We did not use absolute frequencies, as the number of papers including data experts and AI novices differed. Please refer to the table in the appendix ~\ref{app:design_goals} for a numeric overview.
\newpage

%% file: sections/results.tex
\section{Descriptive Results}
\label{sec:descriptive results}
\subsection{Publication Year}
Analyzing the annual count of reviewed papers from 2017 to 2023, the data indicates a general upward trend in the number of papers, beginning at zero in 2016 and 2017 and reaching a peak of 26 papers in 2023. A notable shift in the trend is evident between 2021 and 2022, with a slight decline from 14 to 13 papers published. Then, from 2022 to 2023, a surge follows. This suggests a growing interest or output in the field of human-centered XAI evaluations.
 \subsection{Participants}
Participants of the XAI evaluation were classified as either AI novices or data experts. In some cases, they were specifically declared as data science experts \cite{ghassabiLeveragingKnowledgeGraphs2023}. In other cases, their explicit experience was listed (e.g. publication in the data science field \cite{zhangAdaVisAdaptiveExplainable2023}, degree in data science \cite{weideleAutoDOVizHumanCenteredAutomation2023}). AI novices, could be either complete lay users or domain experts from non-technical fields. For example, domain experts could be biologists \cite{hongVisualizingComparingMachine2023}, healthcare professionals like doctors or nurses \cite{paniguttiCodesignHumancenteredExplainable2023}, or video game players \cite{mirandaInteractiveExplainableCaseBased2021}.
Most papers evaluated their system only with AI novices (n = 55). In 5 papers, systems were only evaluated by data experts. In 5 papers, systems were evaluated by AI novices and data experts.
\section{Synthesis}
\label{sec: synthesis}
\subsection{Taxonomy of XAI System Properties and Evaluation Metrics}
\label{sub-sec: Taxonomy of XAI System Properties and Evaluation Metrics}
We have developed two distinct concept matrices according to the process described in section ~\ref{sec: review method}. The first matrix describes the evaluated XAI system and its properties at the core system and explanation level. The second matrix describes the evaluation metrics at the core system, explanation, and user level. Please refer to the overview of the definition of each coded concept and corresponding examples derived from the literature in tables ~\ref{tab:XAI_properties_concepts} and ~\ref{tab:evaluation_concepts}.

\subsubsection{Evaluated XAI Systems}
Most of the evaluated XAI systems were analysis systems (n=38). These systems provide analysis and evaluation of data regardless of the domain. For example, it could be assessing whether a social media post contains sensitive information \cite{alsulamiExploringUsersPerception2022} (Internet \& Social Media), image classification \cite{chandramouliInteractivePersonalizationClassifiers2023} (Multimodal Data Processing), or fairness judgments \cite{dodgeExplainingModelsEmpirical2019} (Judicial). Analysis systems are not clearly aimed at influencing users' decisions, in contrast to decision support systems. For example, Ibrahim \cite{ibrahimExplanationsImproveQuality2023} used different explanation types in a decision support system to influence participants' risk assessment of a criminal's recidivism (Judicial). On the other hand, recommender systems are mostly aimed at providing recommendations, such as movie recommendations for users \cite{kimImprovedExplanatoryEfficacy2020} (Consumer \& Lifestyle). The presence of different system types in the data shows that XAI is already being applied for various purposes.
\par
Most XAI explanations were graphical (e.g. Graphs \cite{bhattacharyaDirectiveExplanationsMonitoring2023}; n=43) or textual (e.g. written, verbal explanations \cite{alsulamiExploringUsersPerception2022}; n=49). It is important to note, that many XAI explanations were multimodal. Most XAI explanations explained individual outputs locally (n=42). For instance, Chazette \cite{chazetteRequirementsExplanationsQuality2022} showed users the total number of users influencing the calculation of the present route in a navigation task. Local explanations either appeared in addition to a global explanation or on their own. Global explanations refer to explanations that illustrate the system's overall functioning. An example of a global explanation is the work of Cai \cite{caiEffectsExamplebasedExplanations2019}. The authors showed participants the system's training data to explain the classification of an image recognition system. There were several explanatory approaches. The most prominent were feature-based explanations, which also aligns with other literature reviews \cite{sentHumancenteredEvaluationExplainable2024}. Feature-based explanations explain the outcome in terms of the most salient features that lead to its prediction. For example, these could be past sales and market demand leading to a prediction of demand for lemonade \cite{bendavidExplainableAIAdoption2021}, education or marital status for an income prediction \cite{chenUnderstandingRoleHuman2023}, or pixels in a saliency map contributing to the classification of an image \cite{zhangDebiasedCAMMitigateImage2022}. Explanations can be either interactive or static. Interactive explanations could include, for example, clickable explanations that provided more information \cite{wangInterpretableDirectedDiversity2022}, hyperlinks \cite{uenoTrustRelianceConsensusBased2023}, or chatbots that could be prompted \cite{caiEffectsExamplebasedExplanations2019}. Static explanations did not provide opportunities for interaction.
\subsubsection{Evaluation Metrics}
\label{sub-sub-sec: evaluation metrics}
We categorized our core system metrics as follows: \textbf{Affection}, \textbf{Cognition}, \textbf{Usability} and \textbf{Interpretability}. Please note that this framework groups and categorizes evaluation metrics, but does not aim to map relationships between the metrics. Relationships between the metrics could not be identified, as we included metrics purely on the basis of their occurrence. In the following section, we present an overview of the most important metrics for each dimension and propose validated questionnaires. Please refer to table \ref{tab:xai_validated_questionnaires} for a quick overview.
\par
\textbf{Affection} in system metrics refers to users' emotional involvement and feelings 
toward a system during interaction. The most frequently coded affection-related metric was Trust.
Trust is a concept that has been extensively researched in human-machine-interaction (HMI) and
automation contexts. For example, Lee \cite{leeTrustAutomationDesigning2004} published a comprehensive review paper that included an integrative trust model. Their definition of trust was "the attitude that an agent will help achieve an individual’s goals in a situation characterized by uncertainty and vulnerability". Based on our reviewed papers, we defined trust as "Users’ reliance that the system and its output is accurate" (see table \ref{tab:XAI_properties_concepts}).
In this context, trust can be expressed in both attitudes and behaviors. For example, in
our reviewed papers, trust has been measured as an attitude through questionnaires (e.g., \cite{bendavidExplainableAIAdoption2021,bhattacharyaDirectiveExplanationsMonitoring2023}), interviews
(e.g. \cite{hongVisualizingComparingMachine2023}), but also as a behavior (e.g. \cite{ibrahimExplanationsImproveQuality2023,naisehNudgingFrictionApproach2021}). Trust becomes a particularly important metric in XAI systems because the system’s explanations can significantly influence trust \cite{wangRecommendationAgentsElectronic2007} and therefore lead
to system misuse or disuse \cite{leeTrustAutomationDesigning2004}. When analyzing the questionnaires in the reviewed papers, only 4
out of 28 used questionnaires were partially validated (some questions are validated, others are
not) or fully validated (all questions are validated and at most adapted to the usage concept). This
lack of validated questionnaire applications could be addressed by providing validated measures.
For example, the trust between people and automation scale by Jian (TPA) \cite{jianFoundationsEmpiricallyDetermined2000} is widely employed in HMI research. Recently, Hoffman \cite{hoffmanMeasuresExplainableAI2023a} combined several TPA items, along with other items from common trust scales and introduced the trust scale for explainable AI (TXAI). This scale has been further validated and adapted by Perrig \cite{perrigTrustIssuesTrust2023} and presents to the best of our knowledge, the most comprehensive scale for trust measurement in the XAI context. It is important to note that while trust is a commonly measured metric, it does not provide a holistic picture of user attitudes and behaviors toward an AI
system. Therefore, other metrics need to be considered in XAI evaluation in addition to trust.
\par
\textbf{Cognition} encompasses the mental processes users employ to interact with and understand a system. The most commonly measured metric was \textit{understandability}.
In table ~\ref{tab:target_properties_xai}, we referred to understandability as a characteristic of a model that allows a human to understand its function without explaining its internal structure or algorithms. However, understandability as a metric in HMI research evaluates the ease with which software can be understood by individuals \cite{linModelMeasuringSoftware2006}. Similarly, according to the reviewed papers we refered to understandability as "Perceived ease to understand how the system functions". This differs from the initial definition in that it emphasizes the human perception of the system. In the reviewed papers, understandability was primarily assessed through questionnaires (e.g. \cite{kimAlphaDAPRAIbasedExplainable2023}) with only one instance of understandability being examined through an interview \cite{tandonSurfacingAIExplainability2023}. Of the 18 questionnaires measuring understandability in the papers we reviewed, 5 were validated. A useful measure of understandability is the instrument developed by Madsen \cite{madsenMeasuringHumanComputerTrust2000}. Although designed to measure human-computer trust, it provides the subscale "understandability" (e.g. \textit{It is easy to follow what the system does}). 
\par
\textbf{Usability} refers to the extent to which the system allows users to interact with it effectively, intuitively, and smoothly. One usability facet is usefulness, an important metric because it is a critical determinant of system adoption \cite{venkateshTechnologyAcceptanceModel2008}. System adoption is also known as behavioral intention and referred to in our paper as \textit{intention to use}. As with understandability, usefulness was primarily assessed through questionnaires (e.g., \cite{guoEffectRecommendationSource2022}), with a few cases involving interviews (e.g., \cite{spinnerExplAInerVisualAnalytics2019}). Of the 16 questionnaires in the papers we reviewed, 5 were validated. A prominent model that includes perceived usefulness is the "Technology Acceptance Model" (TAM) \cite{venkateshTechnologyAcceptanceModel2008}. This model and its extensions have been repeatedly validated. For example, Kim \cite{kimDifferencesConsumerIntention2019a} measured TAM components, such as perceived ease-of-use, perceived usefulness and intention to use on-demand automobile-related services.
\par
In table ~\ref{tab:target_properties_xai}, we initially referred to \textbf{interpretability} as a passive model feature, providing clarity for humans and \textit{transparency} as the model's provision of information about internal workings and structure. However, after a review of our literature, we revised the definition of transparency from the initial literature to focus on the "perceived clarity of the relationship between feature values and the model's decision." This shift indicates a move from the original definition's focus on merely disclosing technical details of the system's internals to the revised definition's emphasis on the clarity with which users understand how specific features influence model decisions. Furthermore, a high degree of interpretability is often dependent on the presence of explanations (e.g. \cite{guesmiExplainingUserModels2022}), which is an active feature. Therefore, in the context of our results, we discuss interpretability as an active feature, rather than a passive one. We acknowledge however, that this is viewed differently in former reviews (see table \ref{tab:target_properties_xai}). While all reviewed papers provided XAI systems with explanatory features, transparency overall was only measured seven times through unvalidated questionnaires (e.g. \cite{guoEffectRecommendationSource2022}). Hellmann's \cite{hellmannDevelopmentInstrumentMeasuring2022} scales for transparency in recommender systems provide a validated measure with several subscales of the construct. In particular, the subscale "Input" refers to the user's understanding of how features contribute to the model's sections (example item: \textit{"I understood which item characteristics were considered to generate recommendations"}). In addition to measuring metrics related to the core system and how explanations influence these metrics, explanations themselves must be further assessed.
\par
At the level of \textbf{explanations}, \textit{explanation usefulness} and \textit{explanation satisfaction} were identified as relevant metrics. According to the reviewed literature, explanation usefulness refers to the "perceived usefulness of the explanations for understanding the system" and satisfaction to the "satisfaction with the explanations of the system". Surprisingly, only 26 of the 65 papers evaluated the XAI explanation using one of these metrics. Both metrics were assessed exclusively through questionnaires (e.g., \cite{maWhoShouldTrust2023}), with only one exception assessing the explanation usefulness through observation \cite{ghassabiLeveragingKnowledgeGraphs2023}. Out of 17 questionnaires, one assessed explanation usefulness with a validated questionnaire. Of 15 questionnaires measuring explanation satisfaction, one was validated. A validated scale for explanation satisfaction, the "Explanation Satisfaction Scale," was published by Hoffman \cite{hoffmanMeasuresExplainableAI2023a} and, to the best of our knowledge, provides the only validated instrument for measuring explanation satisfaction in an XAI context. Measuring explanation usefulness using a validated questionnaire is somewhat more difficult. Ma \cite{maWhoShouldTrust2023} adapted Laugwitz's \cite{laugwitzConstructionEvaluationUser2008} "User Experience Questionnaire" to measure "perceived helpfulness to decide when to trust the AI". It is generally conceivable that one could use subscales of the UEQ, such as "Perspicuity" (example item: \textit{not understandable - understandable}), or "Dependability" (example item: \textit{obstructive - supportive}) to measure different aspects of explanation usefulness. It should be noted, however, that the UEQ was designed to measure entire products or systems, not just subcomponents such as explanations. 
Besides the components of explanations, user characteristics also need to be taken into account in XAI evaluation. That is, because they can significantly impact their interaction with the system and the feedback they provide.
\par
The most prominent \textbf{user characteristics} in the reviewed literature was \textit{domain expertise}. Domain expertise was assessed in 35 papers. It was frequently measured in questionnaires (e.g. \cite{wangRePromptAutomaticPrompt2023}), but also through preselection (e.g. \cite{ghassabiLeveragingKnowledgeGraphs2023}). The latter means that only individuals with a certain level of domain expertise were selected as participants in the evaluation study, according to their role (e.g. doctors). Out of 19 questionnaires, only one was partially validated. However, to the best of our knowledge, no validated questionnaire measures expertise across domains. This may be because each domain has its specific tasks and requirements. Nevertheless, validated questionnaires to particular domains are available. For instance, if the evaluation study incorporates experts from the healthcare domain, such as nurses, a validated instrument for this domain can be used \cite{gillespieDevelopingModelCompetence2012}. Another way to assess domain expertise might be to measure domain experience (e.g. in years) in a particular position. Hoffmeyer-Zlotnik \cite{hoffmeyer-zlotnikHarmonizedQuestionnaireSociodemographic2018} provided guidelines on socio-demographic questionnaires for orientation. Another way would be to deliberately select participants from one or more specific roles if it can be assumed that different levels of domain expertise are reflected in these positions.
\par
Another area of focus for assessment is the user's \textbf{interaction behavior}. In this context, we define user behavior as actions and interactions that do not directly affect task performance. Some behaviors may also be performance-related, which we have classified as \textit{user task performance}. The assessment of user behavior was primarily based on usage logs, which included data on click behavior (e.g. \cite{alsulamiExploringUsersPerception2022}). In a few cases, user behavior was also assessed through interviews (e.g. \cite{chenUnderstandingRoleHuman2023}) or observation (e.g. \cite{hongVisualizingComparingMachine2023}). Notably, psychophysical measures, such as EEG or eye-tracking, were employed in certain instances (e.g. \cite{liuIncreasingUserTrust2022}). Psychophysical measures can provide an objective perspective on user interactions, but they have inherent limitations, including increased costs due to the necessity of specialized equipment. From this view, usage logs represent a viable, cost-effective option.
While objective metrics are often regarded as superior to subjective metrics \cite{jordanObjectiveSubjectiveMeasurement2024}, a combination of the two, as employed in several reviewed papers, can offer a comprehensive approach \cite{zhouEvaluatingQualityMachine2021, xiongComprehensiveSurveyEvaluation2023}. Overall, selecting an appropriate method for assessing user behavior depends on the context in which it is being conducted and the specific research question being addressed.
\input{Tables/XAI_validated_questionnaires}
\newpage
\subsection{Design Goals}
\label{sub-sec: Design Goals}
\begin{figure}[H]
		\caption{User Groups of XAI with Extended Design Goals adapted from Mohseni \cite{mohseniMultidisciplinarySurveyFramework2021}.}
		\includegraphics[width=14cm]{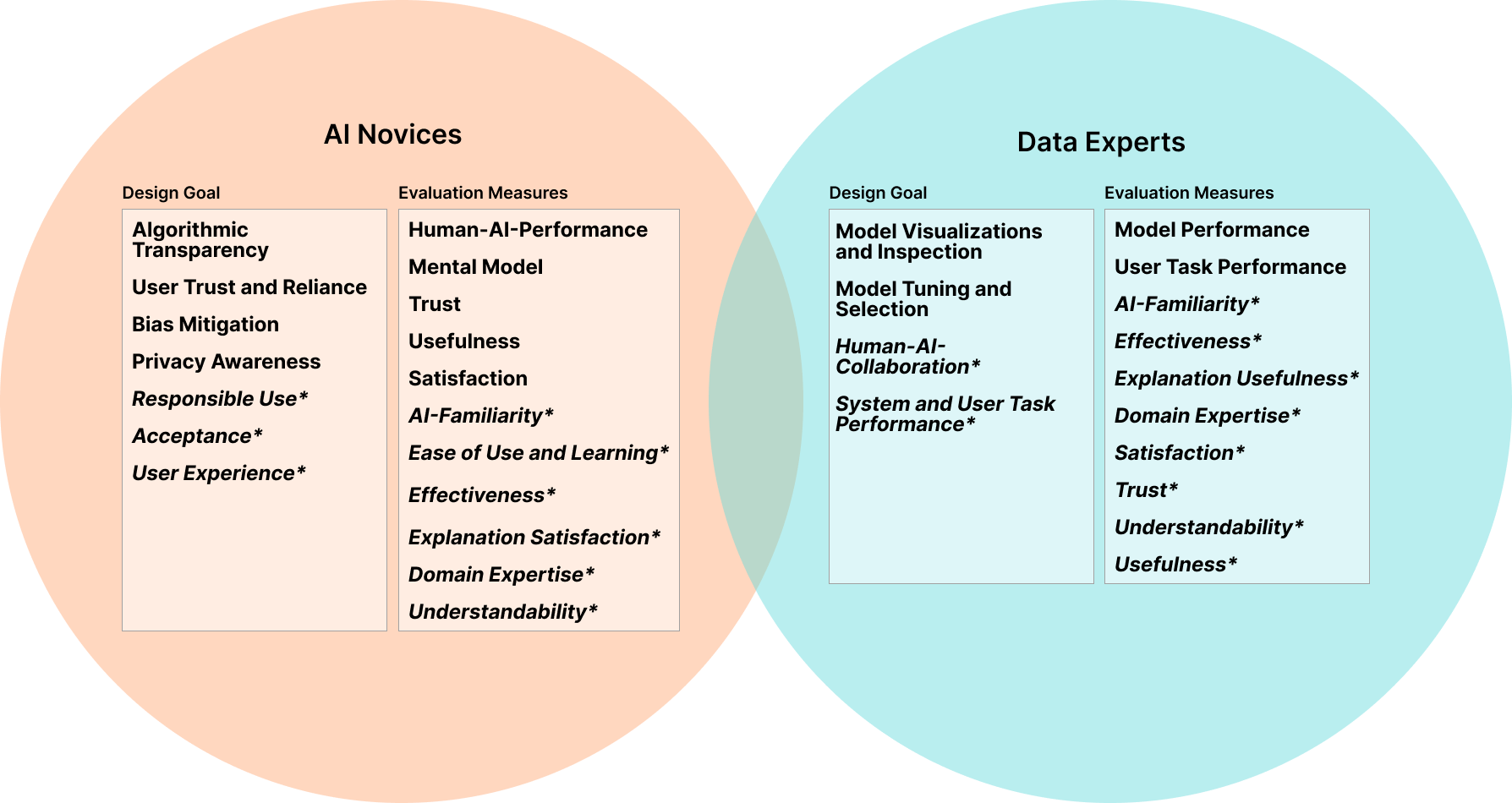}
		\label{fig: user_groups_extended}
            \begin{minipage}{0.85\linewidth}
        \footnotesize \textit{Note:} Design goals and metrics with asterisks in italic were added by the authors of this paper.
    \end{minipage}
	\end{figure}

In this review, we aimed to extend the design goals for AI novices and data experts from the framework of Mohseni \cite{mohseniMultidisciplinarySurveyFramework2021}. For AI novices, we added the design goals of \textit{responsible use}, \textit{acceptance}, and \textit{user experience}. In addition, the design goals for data experts were expanded to include \textit{human-AI collaboration} and \textit{system and user task performance}. Refer to figure ~\ref{fig: user_groups_extended} for an overview.
\subsubsection{AI Novices}
The first additional design goal derived from the literature was to achieve \textbf{responsibility}. The papers refered to this design goal by aiming for \textit{appropriate trust} or \textit{calibrated trust}. This expands the approach of simply increasing \textit{trust} by designing systems in such a way that neither undertrust nor overtrust arises. For example chen \cite{chenUnderstandingRoleHuman2023} argue that the prevention of harmful overreliance is a facilitator of responsible AI design. This can be achieved by transparently disclosing the system's inner workings and its decision making processes. For instance, Bhattacharya \cite{bhattacharyaDirectiveExplanationsMonitoring2023} used various visual and textual explanations to give users an overview of Other system's were aimed to enhance (perceived) \textit{fairness} of model decisions. For instance, Dodge \cite{dodgeExplainingModelsEmpirical2019} investigated the impact of different explanations on users' fairness judgements. In addition Kusuma \cite{kusumaCivilWarTwin2022} highlighted, that AI systems can face a number of challenges, such as privacy, transparency and bias which need to be adressed in system design. Consequently, the design goal of \textbf{responsibility} includes the incorporation of ethical questions, such as inappropriate tust, algorithmic unfairness and bias or privacy into the system design.
\par
The second design goal assigned to AI novices is \textbf{acceptance}. This design goal is closely related to \textit{intention to use}, also refered to as \textit{adoption}. For instance, Ben david \cite{bendavidExplainableAIAdoption2021} investigated the effect of explanation type on readiness to adopt. In their study, a higher readiness to adopt was associated with higher trust in the algorithm and a higher explanation satisfaction. Thus, creating trustworthy systems with favorable explanations can contribute to this goal. The study by Kim \cite{kimAlphaDAPRAIbasedExplainable2023} confirmed that the integration of explanatory features into the system can indeed increase intention to use. Bhattacharya \cite{bhattacharyaDirectiveExplanationsMonitoring2023} further argue that explanations often lack the explanatory depth required by lay-users to accept model decisions. Thus, they combined local and global explanations to provide multiple perspectives and increase user acceptance. Overall, designing for \textbf{acceptance} includes increasing usage intention and system adoption. This can be achieved by providing useful and satisfying explanations.
\par
The final design goal for AI novices was to improve the \textbf{user experience}. This design goal is also related to providing faithful explanations, as misleading explanations can negatively impact the user experience \cite{zhangDebiasedCAMMitigateImage2022}. Different explanation types can also lead to different degrees of user experience. For example, Yang \cite{yangHowVisualExplanations2020} found that image-based explanations lead to a better user experience than charts. Furthermore, algorithmic trade-offs such as the tension between algorithmic fairness and accuracy can influence user experience \cite{yuanContextualizingUserPerceptions2023}. You \cite{youSelfdiagnosisHowChatbotbased2023} also reported that there is a trade-off between the quantity of explanations and user experience. Since long, numerous explanations enhance transparency, they decrease user experience as users need to read or interact with them.
\textit{Usability} is a metric, closely related to user experience. As this is frequently measured, there seems to be an urge to create usable systems. However, as described above, improving user experience involves several trade-offs, such as the accuracy and quantity of explanations.
\subsubsection{Data Experts}
One design goal for data experts is to enhance \textit{human-AI collaboration}. This is closely related to feedback loops in which systems are iteratively refined based on user feedback. For instance, Hanif \cite{hanifEvidenceBasedPipeline2022} developed an explainable dashboard for data scientists to visualize and quantify model behaviors. Users then were able to tune the model using interactive features of the dashboard. Similarly, Spinner \cite{spinnerExplAInerVisualAnalytics2019} let users interactively refine and optimize the models. This process was accompanied by visual feedback, when a refinement was made. Piorkowski \cite{piorkowskiAIMEEExploratoryStudy2023} provided a system which visualizes, explains and provides the possibility to modify model decision boundaries. Further, it served as a tool to facilitate the collbaboration between data scientists and others, such as business stakeholders. Thus, human-AI collaboration consists of interactive \textit{feedback} loops between humans and the AI, but at the same time it can moderate the collaboration between humans with different roles.
\par
The second design goal for data experts aims to enhance \textbf{system task performance}. Zhang \cite{zhangAdaVisAdaptiveExplainable2023} introduced an adaptive and explainable visualization recommendations for tabular data. They evaluated the system \textit{effectiveness} by ranking the correct visualization choices. Further, they gathered qualitative user feedback to estimate the presentation of the recommended visualization. Weidele \cite{weideleAutoDOVizHumanCenteredAutomation2023} argued, that developers need tools to monitor model performance and the execution flow. Thus, they developed a system to automatically find the best machine learning pipeline. \cite{sunDesigningDirectFeedback2023} measured system performance by using computer-centered metrics, such as model accuracy. They also used qualitative user feedback generated in interviews. For instance, they asked for the \textit{effectiveness} of saliency maps in evaluating future model performance. Thus, system performance can be measured by established, computer-centric metrics, but it can also be estimated by collecting user feedback. Refer to appendix \ref{app:design_goals} for a explanations, examples and keywords of the proposed design goals.

\subsection{Overlap}
It is important to consider, that assigned design goals also appeared in the respective other group. There, they were simply not as prevalent. For instance, \textbf{human-AI collaboration} also appeared in systems designed for AI novices. Chen \cite{chenUnderstandingRoleHuman2023} investigated the types of human intuition present in human-AI-decision making with explanations. They refer to human-AI decision making as an AI-assisted process in which the model helps the user to make final judgments or decisions. Another example of human-AI collaboration in AI novices is Hernandez-Bocanegra's \cite{hernandez-bocanegraExplainingRecommendationsConversations2023} system, which allows users to interactively query natural language explanations, based on feedback from a former user study. Compared to human-AI collaboration in data expert systems, AI novices systems aim to facilitate indirect manipulation or assisted decision-making by incorporating feedback. Human-AI collaboration in data expert systems often refers to direct model parameter manipulation \cite{weideleAutoDOVizHumanCenteredAutomation2023, sunDesigningDirectFeedback2023}.
Further, Draws \cite{drawsExplainableCrossTopicStance2023} introduced a system to help users navigate debated topics online while evaluating the model's \textbf{task performance} (text classifications). Similarly, the design goals assigned to AI novices also appeared in systems designated to data experts. For instance, Ghassabi \cite{ghassabiLeveragingKnowledgeGraphs2023} emphasize, that they incorporate explanations to make users understand the process and empower them to make \textit{informed decisions}, which is related to \textbf{responsibility}. The design goal of \textbf{acceptance} did appear in one paper targeted both at AI novices and data experts but in no paper targeted solely at data experts. Regarding \textbf{user experience}, Weidele \cite{weideleAutoDOVizHumanCenteredAutomation2023} interpreted this term in a broader sense. They referred to the system design, its implementation and its \textit{usability}. Thus, the design goal of \textbf{user experience} is incorporated but referred to in a more generic way, rather than directly focusing on improving user flow and ease of use.

%% file: Tables/XAI_validated_questionnaires.tex
\renewcommand{\arraystretch}{1.5}

\begin{table*}[ht]
\centering
\caption{XAI Evaluation Metrics and Proposed Validated Questionnaires}
\label{tab:xai_validated_questionnaires}
\small

\begin{tabularx}{\textwidth}{%
    >{\bfseries\raggedright\arraybackslash}p{2.4cm}
    >{\raggedright\arraybackslash}p{3.0cm}
    >{\raggedright\arraybackslash}X
}
\toprule
\textbf{Category} &
\textbf{Metric} &
\textbf{Proposed Validated Questionnaire(s)} \\
\midrule

\multirow[t]{2}{2.4cm}{\textbf{Affection}} 
& Trust 
& \textbf{TXAI} Trust Scale for Explainable AI \cite{hoffmanMeasuresExplainableAI2023a}, adapted by Perrig \cite{perrigTrustIssuesTrust2023}. \\

\cmidrule(l){2-3}
& Transparency 
& No fully validated scale identified; transparency in recommender systems subscales by Hellmann \cite{hellmannDevelopmentInstrumentMeasuring2022}. \\

\midrule

\multirow[t]{2}{2.4cm}{\textbf{Cognition}} 
& Understandability 
& \textbf{Human-Computer Trust Scale} by \cite{madsenMeasuringHumanComputerTrust2000}, subscale \emph{Understandability}; e.g., \emph{``It is easy to follow what the system does.''} \\

\cmidrule(l){2-3}
& Interpretability 
& No fully validated scale identified; transparency in recommender systems subscales by Hellmann \cite{hellmannDevelopmentInstrumentMeasuring2022}. \\

\midrule

\multirow[t]{2}{2.4cm}{\textbf{Usability}} 
& Usefulness 
& \textbf{TAM} Technology Acceptance Model \cite{davisPerceivedUsefulnessPerceived1989}, validated by Kim \cite{kimDifferencesConsumerIntention2019a}; subscales: \emph{Perceived Usefulness} and \emph{Perceived Ease-of-Use}. \\

\cmidrule(l){2-3}
& Intention to Use 
& \textbf{TAM} Technology Acceptance Model \cite{davisPerceivedUsefulnessPerceived1989}, validated by Kim \cite{kimDifferencesConsumerIntention2019a}; subscale: \emph{Intention to Use}. \\

\midrule

\multirow[t]{2}{2.4cm}{\textbf{Explanation Quality}} 
& Explanation Satisfaction 
& \textbf{Explanation Satisfaction Scale} \cite{hoffmanMeasuresExplainableAI2023a}. \\

\cmidrule(l){2-3}
& Explanation Usefulness 
& \textbf{UEQ} User Experience Questionnaire \cite{laugwitzConstructionEvaluationUser2008}, initially designed to measure whole-system user experience. \\

\midrule

\multirow[t]{2}{2.4cm}{\textbf{User Characteristics}} 
& Domain Expertise 
& No cross-domain validated questionnaire; domain-specific instruments are available, e.g., \cite{gillespieDevelopingModelCompetence2012}. Alternatively, domain expertise can be assessed via years of experience or role-based preselection \cite{gillespieDevelopingModelCompetence2012,hoffmeyer-zlotnikHarmonizedQuestionnaireSociodemographic2018}. \\

\cmidrule(l){2-3}
& User Behavior 
& No questionnaire; assessed via usage logs, observation, interviews, or psychophysical measures, e.g., EEG or eye-tracking. \\

\bottomrule
\end{tabularx}
\begin{minipage}{0.95\linewidth}
\footnotesize \textit{Note:} This table presents a curated overview of validated questionnaires that assess the constructs identified through the manual literature review.
\end{minipage}
\end{table*}

%% file: sections/discussion.tex
\section{Discussion}
\label{sec:discussion}
Our review was guided by two questions: how existing XAI evaluation metrics can be integrated to enable human-centered evaluation (RQ1), and which design goals can be derived from the evaluation literature (RQ2). For RQ1, the synthesis in Section 5.1 produced a three-level framework : \textit{core system}, \textit{explanation}, and \textit{user}.  Within which core-system metrics group into \textbf{affection}, \textbf{cognition}, \textbf{usability}, and \textbf{interpretability}. For RQ2, Section 5.2 extended Mohseni's  \cite{mohseniMultidisciplinarySurveyFramework2021} goals with \textit{responsibility}, \textit{acceptance}, and \textit{user experience} for AI novices, and \textit{human-AI collaboration} and \textit{system/user task performance} for data experts. While answering these questions we observed recurring shortcomings: sparse questionnaire validation, incomplete method reporting, narrow metric coverage, under-evaluation of explanations, and neglect of behavioral intention. The following five guidelines respond to these shortcomings; we then show how the design goals (RQ2) govern which framework metrics (RQ1) an evaluation should prioritize.
\subsection{Guidelines and Recommendations}
\subsubsection{Guideline 1: Validation and Standardization}
We analyzed the share of validated questionnaires in popular XAI evaluation metrics. Our research indicates that validated questionnaires are used sparingly at present. This is also in line with the findings of related reviews (e.g. \cite{sentHumancenteredEvaluationExplainable2024, naveedOverviewEmpiricalEvaluation2024}). The absence of questionnaire validation is a significant issue, as it leaves open the possibility that the intended construct is not accurately measured \cite{taherdoostValidityReliabilityResearch2016}. This can result in flawed interpretations and design decisions. Furthermore, this can hinder the comparability of user evaluation studies. We therefore advise to use validated questionnaires, such as the explanation satisfaction scale by Hoffman \cite{hoffmanMetricsExplainableAI2019a}. Another recommendation is the standardization of the user testing. This can involve standardizing qualitative user interviews by utilizing pre-defined questions, as well as standardizing the whole study procedure. Standardizing user testing enhances the comparability of results by ensuring that users receive consistent instructions and questions and are subject to comparable procedures. Furthermore, it can facilitate reporting of the study methodology, which is another area of improvement in the current XAI evaluation literature. Selecting validated instruments should be guided by the design goal under evaluation: for instance, the TXAI trust scale supports assessing \textit{responsibility} in AI novices, while the TAM intention-to-use sub-scale supports evaluating \textit{acceptance}.
\subsubsection{Guideline 2: Presentation of Study Methodology}
One of our exclusion criteria for papers in the filtering process was the lack of a clear description of methods and related constructs. That was because we were not able to trace \textit{what was measured} and \textit{how it was measured} in some of the papers. We therefore had to exclude them from our analysis. This is a problem not only for systematic reviews like ours but also for other XAI researchers. Clearly presenting procedures can provide transparency about the research that has been done. This helps to establish whether or not the process was appropriate. It can also open up the possibility of related studies testing the same relationships in different contexts and adopting the same procedure. We therefore recommend rigorous documentation from study planning to data analysis. This is useful to gather the necessary information for the paper's methods section. For a guide to write the methods section, please see Kallet \cite{kalletHowWriteMethods2004}.  This guideline is not linked to a particular design goal but applies across all of them; reporting should nonetheless state which design goal an evaluation targets, making the rationale behind the selected metrics transparent.
\subsubsection{Guideline 3: Holistic Evaluation}
One of our key findings was that certain metrics (e.g. trust) were measured much more frequently than others (also see Naveed \cite{naveedOverviewEmpiricalEvaluation2024}). This can be because poorly calibrated trust can lead to the misuse or disuse of a system \cite{leeTrustAutomationDesigning2004}, which – especially in risk-associated domains – can have serious consequences. It may also be due to a growing interest in trust in the wider literature \cite{benkTwentyfourYearsEmpirical2024}, which has led to a need to incorporate this metric into evaluation studies. While we acknowledge the importance of trust as a metric in XAI evaluation research, we believe there is a need for a more holistic approach to evaluating such systems. We identified four core areas for consideration: affection, cognition, usability, and interpretability. In addition, we inductively derived further dimensions: the explanation, user characteristics, and user interaction behavior. We argue that each of these dimensions should be considered when evaluating XAI systems, as they provide different perspectives on the system. Furthermore, we believe the metrics used must be strongly aligned with the research questions. When selecting metrics, it is important to establish \textit{what} is being measured and most importantly \textit{why} before engaging in the question of \textit{how} it is measured.  Which dimensions to emphasize depends on the design goal: novice goals like responsibility draw on affection and interpretability, expert-facing goals like system task performance on usability and performance.
\subsubsection{Guideline 4: Evaluation of Explanations}
In our analyses, we identified a lack of evaluation of explanations in the reviewed literature. While all of the reviewed papers incorporated an XAI system, less than half of them evaluated the explanation or at least the overall transparency of the system. However, failure to evaluate this aspect of the system results in an incomplete evaluation and possibly problematic interpretations. For instance, an XAI system may foster positive user attitudes such as satisfaction and behaviors like user task performance, yet users may remain unaware of how the system operates. This could lead to an unnoticed redundancy of the implemented explanations. We posit that all XAI evaluation studies should integrate measures related to explanations. To gain insight into the key elements of XAI explanations, we suggest utilizing the framework outlined by Kim \cite{sentHumancenteredEvaluationExplainable2024}. Evaluating explanations directly supports the AI-novice design goals of acceptance and user experience, since both depend on explanation usefulness and explanation satisfaction as their underlying metrics.
\subsubsection{Guideline 5: Consideration of Behavioral Intentions}
As outlined in the technology acceptance model \cite{venkateshTechnologyAcceptanceModel2008}, behavioral intention is a direct predictor of use behavior. However, our analysis revealed that less than a quarter of the reviewed papers considered behavioral intention in the form of \textit{intention to use}. In controlled experiments, where the objective is to investigate causal relationships between variables, this is a less pressing concern. However, in the context of applied research with real systems, this is a significant shortcoming. That is because such systems may undergo further development to deploy them to users. Even though other metrics such as \textit{perceived usefulness} are also related to system adoption, they are rather predictors of \textit{intention to use} than direct predictors of \textit{usage behavior} \cite{venkateshTechnologyAcceptanceModel2008}. It is therefore recommended to measure intention to use (as a direct predictor of usage behavior) in all applied XAI user studies.  This guideline is tied most directly to the design goal of acceptance, for which intention to use is the central evaluation metric.
\subsubsection{Application of Design Goals}
We introduced an extension of the design goals for AI novices and data experts by Mohseni \cite{mohseniMultidisciplinarySurveyFramework2021}. In general, the design goals for novices were primarily focused on ethical considerations (e.g. appropriate tust), while those of data experts were centered on the broader collaboration between humans and AI, as well as the resulting performance (e.g. system task performance). When conducting XAI evaluation studies, it is important to implement these goals. To this end, we recommend the following procedure: 
\begin{enumerate}
    \item \textbf{Identify the user group}: It is essential to identify the XAI user group before planning the study. This should ideally occur after human-centered requirement analysis and before developing the system \cite{rossonChapter2Analyzing2002}.
    \item \textbf{Select Design Goals}: Once the relevant XAI user group has been identified, a selection of appropriate design goals should be made. It is preferable to address as many design goals as possible related to this user group, but for practical reasons, a focus should be set. This is dependent on the research context, for example, technical feasibility.
    \item \textbf{Implementation of Design Goals in System Development}: Measures to achieve the design goals need to be implemented in system development e.g. information about system limitations to achieve responsible system use.
    \item \textbf{Development of Research Questions}: The formulation of appropriate research questions is based on the selected design goals.
    \item \textbf{Selection of Evaluation Metrics}: It is essential to select appropriate metrics to answer the formulated research questions e.g. intention to use for the design goal of acceptance.
    \item \textbf{Design Goal Validation}: Following the completion of user studies, data analysis on corresponding evaluation metrics is conducted to validate the design goals.
\end{enumerate}
Adherence to these guidelines facilitates the achievement of the recommended design goals. We need to recognize, however, that these guides represent a simplification of the actual development process. The proposed process assumes a rigid, linear progression, which may not align with the iterative nature of system development in the real world.  
\subsection{Future Literature Trends}
This review includes literature published up to and including 2023, reflecting the temporal scope of the composite search and analysis conducted in 2024. To investigate the relevance of our framework, we looked at more current XAI user studies. We reused our search query in the same databases (ACM and IEEE Explore) and searched for literature from the years 2024 and 2025. For this small extension, we applied our taxonomy and design goals to the studies appearing among the ten initial search results for the selected search string. \par
Overall, we found that our proposed framework was still relevant in the current research. For instance, Zhao \cite{zhaoDomainExperienceExpertise2025} investigated domain experts with varying levels of domain expertise regarding their decision-making processes in an AI-assisted task. They assessed diverse metrics, found in our framework, such as explanation usefulness, trust and understandability. Furthermore, The authors investigated the role of users’ domain-related characteristics in XAI, specifically their level of domain expertise. One of their primary design goals was \textbf{acceptance}.  The paper discusses acceptance not as a uniform outcome but as being shaped by the interplay of domain expertise and practical experience, arguing that XAI research should move beyond a simple expert or novice binary since practitioners with similar "expert" status can accept explanations for very different reasons and with different consequences. Langerak \cite{langerakXAIUIUserBeliefDriven2025} evaluated a recipe app. They did so by including context-related variables, such as daytime or the existence of deadlines, and varying the expertise levels of their users. Ultimately, they could produce explanations that were adapted to these user variables. This illustrated a system property, which also occurred in other literature: \textit{contextual awareness}. By that, we refer to the ability of the system to adapt to different situations and/ or users. In the study, the authors used explanations that were created adaptively to the users' characteristics to enhance their experience. Thus, designing contextually aware systems can be understood as a mean to inform the design goal of \textbf{user experience}. Bucur \cite{bucurInteractiveEvolutionaryOptimization2025} incorporated contextual awareness by providing  a variety of accessible explanations for lay users. For evaluation they used various metrics, present in our framework like explanation usefulness, understandability, ease of use, and learning.
Abhiram \cite{abhiramHumanCentricEvaluationXAI2025} applied several XAI methods in a sentiment analysis of Amazon reviews. Their goal was to incorporate \textbf{acceptance}, addressed by fostering the system's adoption. Lastly, Szymanski \cite{szymanskiGranularFeedbackLeveraging2025} let teachers assess AI-generated questions for students. \textbf{User experience} was addressed by iteratively designing and testing four feedback interfaces with real teachers, using their preferences and think-aloud feedback to derive concrete usability design goals. \par
Besides design goals and metrics, we also checked for user groups in the literature sample.
In the initial literature we did not find any user studies conducted with AI experts. Since our search query could have been too restrictive in this regard, we searched the previously mentioned databases again, replacing the block ("user" OR "non-expert user" OR "domain expert" OR "lay user") with "AI expert". However, we still did not find any papers involving AI experts as subjects in XAI user studies. This suggests that AI experts might be underrepresented in XAI evaluation research. Despite their technical focus, they should be included in future research. XAI methods developed to facilitate model debugging and improve model interpretability still need to be understood by AI experts and used in the right application context. Further, for future search attempts, it may be helpful to adapt the search query by including several synonyms of "AI expert" to ensure that nothing is missed.\par
Although we identified new user studies within the existing corpus of XAI literature, the absence of human-centred evaluation remains unchanged. Suh \cite{suhFewer1Explainable2025} found that only 1\% of XAI-related papers evaluate their systems with human users. This highlights a gap that was already evident in earlier literature, which motivated our literature review. It also underscores the importance of taxonomies and frameworks aiming to support XAI developers in human-centered system evaluation.\par
Our framework captures essential design goals and evaluation metrics in XAI user studies. After analyzing a sample of current literature, we find that our framework provides a solid groundwork to design and evaluate XAI systems. Further, we anticipate that future research will focus on developing adaptive XAI systems that change based on the situation or user. Our framework's modular structure allows for further extensions in future research, which is valuable in a fast-paced dynamic research field. Current shortcomings in XAI evaluation research further highlight the relevance of our work.
\subsection{Possible Framework Extensions}
Our suggested XAI evaluation framework still has the potential for further expansion. For instance, Kim \cite{sentHumancenteredEvaluationExplainable2024} provided a framework built on the question of \textit{"What makes explanations meaningful?"}. Consequently, they devised 11 significant characteristics for XAI explanations. However, our results differ from theirs because our framework only includes explicitly assessed measures, whereas theirs includes aspects of explanations that users have (implicitly) wished for. For this reason, we believe that a synthesis could enhance our framework. This would entail a synchronization between the desirable aspects of a system and its explanations and their actual implementation in reality.
Additionally, a scoping review was conducted by Naveed \cite{naveedOverviewEmpiricalEvaluation2024} to analyze trends in the evaluation of XAI. The results included two concept matrices. The first matrix addressed the properties of the studies, such as the objective, scope, and procedure. The second matrix presented the used evaluation measures in the studies. While the latter is similar to our concept matrix, which addresses XAI evaluation measures, the first one could provide an extension when analyzing XAI evaluation studies. However, we additionally addressed the \textit{context} of the system, such as the system type and the explanation type, which was not addressed by Naveed \cite{naveedOverviewEmpiricalEvaluation2024}. Further, we distinguished between the core system, the explanation and the user.
It would be beneficial for future studies to focus on a synthesis of existing XAI evaluation frameworks. This would allow for a more comprehensive approach that incorporates various perspectives. Moreover, future research should concentrate on the creation of validated XAI evaluation questionnaires based on identified concepts of the synthesized frameworks. This could result in greater utilization of validated measurement instruments in XAI research and therefore enhance validity and standardization. Regarding the design goals, we identified them based on studies from different domains. It would be beneficial to conduct domain-specific studies to determine which design goals are important in which domain.
\subsection{Limitations of the Framework and Design Goals}
The framework and guidelines proposed in this paper have limitations. First, our approach only reflects the current state of the art in XAI evaluation practice. We included concepts that appeared in at least five papers. This means that we included design goals and metrics that were incorporated by at least five developer teams. However, they do not necessarily align with pre-defined, empirically gathered user requirements. This potential lack of user involvement in early design stages could result in the integration of evaluation metrics and design goals that appear relevant to developers but are not as crucial for users. Nevertheless, keeping current evaluation trends in focus remains a sound starting point. That is, because developer teams incorporate research questions relevant to further system development, and concepts advised by the literature into their evaluation studies. Furthermore, it facilitates the provision of concepts that can be explicitly measured because they are already applied. Lastly, it helps to identify significant shortcomings in current practices.
\par
A second limitation relates to terminology. We found that many different terms were used for similar concepts, e.g. \textit{explanation usefulness} and \textit{explanation helpfulness} or \textit{trust} and \textit{reliance}. However, to summarize, we have grouped them under one umbrella term. This may have resulted in information loss, as the concepts are similar but differ in some details. Furthermore, there is some overlap in the content of the identified concepts. For example, the perceived \textit{effectiveness} of a system is likely to depend on its perceived \textit{efficiency}. However, information regarding the (causal) relationship between these concepts is not provided by the data and therefore not addressed in our review. The limitation of concept overlap also applies to the design goals. For instance, the realization of the goal of \textit{human-AI-collaboration} may also be associated with \textit{system and user task performance}. This is because systems that facilitate effective collaboration between humans and AI could also result in enhanced system and user task performance. Despite the overlaps, we maintain that our concepts differ in certain key aspects, which justifies their distinction.
\par
Furthermore, it should be noted that design goals were assigned to a user group if they appeared in the paper with this user group as the target audience. Consequently, some design goals appeared in both user groups. So, even design goals that are primarily associated with one user group might also be relevant for another user group. Therefore, the categorization should be regarded as an orientation and be adapted to the specific use case.
\par
Lastly, we applied a narrowed search string across two databases. This search string was refined and optimized through several preliminary search attempts. Broader databases were excluded because they generated a large number of generic results and duplicates, which exceeded the scope of our available resources. Although we carefully selected two leading computer science databases, it is possible that relevant literature may have been missed.
\par
Despite these limitations, our framework and design goals can provide researchers with a useful orientation for developing and evaluating XAI systems from a holistic, human-centered perspective, incorporating relevant user groups.

%% file: sections/conclusion.tex
\section{Conclusion}
We conducted a composite literature review of 65 papers that assess XAI systems with human users. Our analysis revealed a multitude of metrics commonly utilized in XAI evaluation research. To facilitate the classification of these components, we have proposed a framework that groups them into three categories, based on which component is evaluated: the \textbf{core system}, the \textbf{explanation}, and the \textbf{user}. Additionally, we identified that the core system was evaluated in the dimensions of users' \textit{affection} towards the system, their \textit{cognitive processes}, its \textit{usability}, and \textit{interpretability}.
Furthermore, we identified relevant design goals for \textbf{AI novices} and \textbf{data experts}: \textit{responsibility}, \textit{acceptance}, \textit{usability} as well as \textit{human-AI-collaboration} and \textit{system task performance}. Lastly, we have provided guidelines for conducting XAI evaluations based on the most commonly identified shortcomings in the literature. Our framework provides a holistic overview of evaluation metrics and XAI systems rather than focusing on specific features. Furthermore, we expand upon the design goals outlined in previous studies by introducing additional goals to be considered during the development process. It is not intended to offer detailed recommendations for user interface and interaction design, such as XAI explanation techniques or design components to include in the user interface. Furthermore, it is designed with a \textbf{human-centered} approach to XAI explanations, rather than a \textbf{computer-centered} one.

%% file: sections/appendices.tex
\appendix
\begin{landscape}
\section*{Appendix}
\label{app:prior_work}
\section{Prior Work}
\input{Tables/Taxonomies_XAI_Evaluation}
\input{Tables/distinction_literature_review}
\end{landscape}
\label{app:excluded}
\section{Excluded Concepts}
\input{Tables/Excluded}
\newpage
\label{app:matrix}
\section{Concept Definitions}
\input{Tables/XAI_Properties_Concepts_Explanations}
\input{Tables/Evaluation_Concepts_Explanations}
\section{Concept Matrices}
\input{Tables/Concept_Matrix_Properties}
\input{Tables/Concept_Matrix_Evaluation}
\input{Tables/System_Properties_Condensed}
\newpage
\section{Design Goals}
\label{app:design_goals}
\input{Tables/Design_Goal_Concepts}
\input{Tables/Design_Goals_User_Groups_Frequencies}

%% file: Tables/Taxonomies_XAI_Evaluation.tex
\begin{table}[ht]
\caption{Existing Taxonomies in XAI Evaluation}
\label{tab:taxonomies_in_XAI_evaluation}
\centering
\begin{tabular}{p{0.3\linewidth}p{0.65\linewidth}}
\hline
\textbf{Evaluation Aspect} & \textbf{Description} \\
\hline
XAI Evaluation Approach \cite{hoffmanMetricsExplainableAI2019a, lopesXAISystemsEvaluation2022} & 
\begin{itemize}
    \item Decontextualized without users (\textcolor{gray}{computer-centered})
    \item In-context of use, with users (\textbf{human-centered})
\end{itemize} \\
XAI Evaluation Levels \cite{sentHumancenteredEvaluationExplainable2024} & 
\begin{itemize}
    \item Evaluation of the \textbf{quality of explanations}
    \item Evaluation of the \textbf{contribution of explanations to user experience}
\end{itemize} \\
\hline
User Types and Tasks & 
\begin{itemize}
    \item \textcolor{gray}{Functionality-grounded}: evaluations that do not involve users and use proxy tasks \cite{doshi-velezRigorousScienceInterpretable2017}
    \item \textbf{Human-grounded}: evaluations involving laypeople on simplified tasks \cite{doshi-velezRigorousScienceInterpretable2017}, also known as \textbf{concept-driven} evaluations \cite{naveedOverviewEmpiricalEvaluation2024}
    \item \textbf{Application-grounded}: evaluations with domain experts and real tasks, also referred to as \textbf{domain-driven} evaluation \cite{naveedOverviewEmpiricalEvaluation2024}
\end{itemize} \\
User Types \cite{mohseniMultidisciplinarySurveyFramework2021} & 
\begin{itemize}
    \item \textcolor{gray}{AI experts} (not present in our literature)
    \item \textbf{Data experts}
    \item \textbf{AI Novices}
\end{itemize}\\
\hline
Evaluation Scenarios \cite{naveedOverviewEmpiricalEvaluation2024} & 
\begin{itemize}
    \item \textbf{Real-world scenarios} with highly critical impact
    \item \textbf{Illustrative scenarios} with less critical impact
\end{itemize} \\
\hline
Metric Types & 
\begin{itemize}
    \item \textbf {quantitative data type} vs. \textbf{qualitative data type}\cite{nautaAnecdotalEvidenceQuantitative2023, viloneNotionsExplainabilityEvaluation2021}
    \item \textbf{subjective data perspective} vs. \textbf{objective data perspective}\cite{zhouEvaluatingQualityMachine2021}
\end{itemize} \\
\hline
\end{tabular}
\label{tab:xai_evaluation}
\caption*{\small \textit{Note:} Aspects in grey are not the subject of this review paper. Aspects in bold are subject of this review paper.}
\end{table}

%% file: Tables/distinction_literature_review.tex
\begin{longtable}{m{0.1\linewidth}m{0.1\linewidth}
>{\raggedright\arraybackslash}m{0.2\linewidth}
>{\raggedright\arraybackslash}m{0.1\linewidth}
>{\raggedright\arraybackslash}m{0.1\linewidth}
>{\raggedright\arraybackslash}m{0.2\linewidth}
>{\raggedright\arraybackslash}m{0.1\linewidth}
}
\caption{Overview of Literature Reviews}
\label{tab:distinction_literature_review}\\
\hline
\textbf{Authors} & \textbf{Publication Year} & \textbf{Research Questions} & \textbf{Review Method} & \textbf{Resulting Artefacts} & \textbf{Distinction from Our Review} & Perspective \\
\hline
\cite{doshi-velezRigorousScienceInterpretable2017} & 2018 & Not explicitly stated & Not explicitly stated (own method) & Taxonomy & 
\begin{itemize}
    \item Methodological approach (unknown review method)
    \item Focus (overall evaluation approaches)
\end{itemize}
& human-centered
\\
\hline
\cite{mohseniMultidisciplinarySurveyFramework2021} & 2021 & Not explicitly stated & Not explicitly stated (own method) & Concept Matrices, Design Goals & 
\begin{itemize}
    \item Methodological approach (unknown review method)
    \item Foundation of our work
\end{itemize}
& computer-centered \& human-centered
\\
\hline
\cite{viloneNotionsExplainabilityEvaluation2021} & 2021 & Not explicitly stated & Not explicitly stated & Tables & 
\begin{itemize}
    \item Methodological approach (unknown review method)
\end{itemize} &  computer-centered
\\
\hline
\cite{zhouEvaluatingQualityMachine2021} & 2021 & Not explicitly stated & Not explicitly stated & Taxonomy, Tables & 
\begin{itemize}
    \item Methodological approach (unknown review method)
    \item Focus (overall evaluation approaches)
\end{itemize}
&  computer-centered \& human-centered
\\
\hline
\cite{lopesXAISystemsEvaluation2022} & 2022 & Not explicitly stated & Not explicitly stated (own method) & Tables &
\begin{itemize}
    \item Methodological approach (unknown review method)
\end{itemize} 
& computer-centered \& human-centered
\\
\hline
\cite{nautaAnecdotalEvidenceQuantitative2023} & 2023 & Not explicitly stated & Not explicitly stated (own method) & 
Co-12 explanation properties, Tables &
\begin{itemize}
    \item Perspective (computer-centered and human-centered evaluation
    \item Focus (co-12 desired explanation qualities)
\end{itemize} & mostly computer-centered
\\
\hline
\cite{naveedOverviewEmpiricalEvaluation2024} & 2024 & 
\begin{itemize}
\item “What are the common practices in terms of patterns and essential elements in empirical evaluations of AI explanations?”
\item “What pitfalls, but also best practices, standards, and benchmarks, should be established for empirical evaluations of AI explanations?”
\end{itemize} 
& Scoping review; unsystematic & Concept Matrices, evaluation guidelines & 
\begin{itemize}
    \item Methodological approach (unstructured)
    \item Focus (solely on metrics)
\end{itemize} & human-centered
\\
\hline
\cite{sentHumancenteredEvaluationExplainable2024} & 2024 &\begin{itemize}
    \item “How is the meaningfulness of XAI explanations evaluated in user studies?”
\end{itemize}
 & PRISMA; systematic & Taxonomy & 
 \begin{itemize}
     \item  Focus (meaningfulness of explanations)
 \end{itemize} & human-centered
 \\
\hline
Our review & - &\begin{itemize}
\item “How can existing XAI evaluation metrics be integrated to enable a human-centered evaluation of XAI systems?”
\item “Which design goals can be derived from existing XAI evaluation literature?”
\end{itemize} 
& Composite literature review method; structured & 
Concept Matrices, Design Goals, evaluation and design guidelines, taxonomy & - &\\ 
\hline
\end{longtable}

%% file: Tables/Excluded.tex
\begin{longtable}{p{0.5\linewidth}p{0.5\linewidth}}
\hline
\textbf{Concept} & \textbf{Papers} \\
\hline
\endfirsthead

\hline
\textbf{Concept} & \textbf{Papers} \\
\hline
\endhead

\hline
\endfoot

\hline
\endlastfoot

Acceptance of estimation & \cite{maWhoShouldTrust2023} \\
Accessibility & \cite{sabuncuogluDevelopingMultimodalClassroom2023} \\
Accountability & \cite{ibrahimExplanationsImproveQuality2023} \\
Actionability & \cite{bhattacharyaDirectiveExplanationsMonitoring2023} \\
Adaptability & \cite{zhangAdaVisAdaptiveExplainable2023} \\
Aesthetics & \cite{sabuncuogluDevelopingMultimodalClassroom2023} \\
Affective state & \cite{kimImprovedExplanatoryEfficacy2020} \\
Affect & \cite{raymondCultureBasedExplainableHumanAgent2020} \\
Agency & \cite{guerdanAffectiveXAIFacial2021} \\
Agreement & \cite{raymondCultureBasedExplainableHumanAgent2020} \\
Attitude & \cite{weideleAutoDOVizHumanCenteredAutomation2023} \\
Attitudes on debated topics & \cite{drawsExplainableCrossTopicStance2023} \\
Auditory Explanation & \cite{danryWearableReasonerEnhanced2020} \\
Awareness & \cite{tsaiExploringPromotingDiagnostic2021} \\
Benevolence & \cite{guoEffectRecommendationSource2022}, \cite{caiEffectsExamplebasedExplanations2019} \\
Big Five Personality Traits & \cite{tutulInvestigatingTrustHumanMachine2021} \\
Broad interests & \cite{alsulamiExploringUsersPerception2022} \\
Capability & \cite{caiEffectsExamplebasedExplanations2019} \\
Challenge & \cite{raymondCultureBasedExplainableHumanAgent2020}, \cite{guerdanAffectiveXAIFacial2021} \\
Comfortableness & \cite{kimPredictionRetrospectionIntegrating2022} \\
Competence & \cite{guerdanAffectiveXAIFacial2021} \\
Computation Diversity & \cite{wangInterpretableDirectedDiversity2022} \\
Confidence in recommendation & \cite{tandonSurfacingAIExplainability2023} \\
Consistency & \cite{kimPredictionRetrospectionIntegrating2022} \\
Control & \cite{leeLIMEADEAIExplanations2023} \\
Desire for emotional support & \cite{youSelfdiagnosisHowChatbotbased2023} \\
Demand for explanations & \cite{chazetteRequirementsExplanationsQuality2022} \\
Emotional support & \cite{youSelfdiagnosisHowChatbotbased2023} \\
Estimation of AI’s confidence & \cite{mishraWhyWhyNot2022} \\
Evaluation strategy & \cite{liuIncreasingUserTrust2022} \\
Explanation agreement & \cite{ghaiExplainableActiveLearning2021} \\
Explanation clarity & \cite{hernandez-bocanegraExplainingRecommendationsConversations2023} \\
Explanation completeness & \cite{ghassabiLeveragingKnowledgeGraphs2023}, \cite{hernandez-bocanegraExplainingRecommendationsConversations2023} \\
Explanation soundness & \cite{ghassabiLeveragingKnowledgeGraphs2023} \\
Explanation transparency & \cite{chazetteRequirementsExplanationsQuality2022} \\
Explainability & \cite{chandramouliInteractivePersonalizationClassifiers2023}, \cite{guoEffectRecommendationSource2022}, \cite{kimAlphaDAPRAIbasedExplainable2023} \\
Expected aesthetics score & \cite{ohUnderstandingHowPeople2020} \\
Fairness & \cite{dodgeExplainingModelsEmpirical2019}, \cite{ibrahimExplanationsImproveQuality2023} \\
Friendliness & \cite{kimPredictionRetrospectionIntegrating2022} \\
Fulfillingness & \cite{kimPredictionRetrospectionIntegrating2022} \\
Fun & \cite{kimPredictionRetrospectionIntegrating2022} \\
Facilitating condition & \cite{paniguttiCodesignHumancenteredExplainable2023} \\
Goal congruence & \cite{guerdanAffectiveXAIFacial2021} \\
Human-likeness & \cite{ehsanAutomatedRationaleGeneration2019}, \cite{youSelfdiagnosisHowChatbotbased2023} \\
Image & \cite{paniguttiCodesignHumancenteredExplainable2023} \\
Involvement & \cite{paniguttiCodesignHumancenteredExplainable2023} \\
Integrity belief & \cite{guoEffectRecommendationSource2022} \\
Interface Evaluation & \cite{liuIncreasingUserTrust2022} \\
Likeability & \cite{youSelfdiagnosisHowChatbotbased2023} \\
Model selection & \cite{yuKeepingDesignersLoop2020} \\
Operability & \cite{sabuncuogluDevelopingMultimodalClassroom2023} \\
Perceived error rate & \cite{kimBubbleuExploringAugmented2023} \\
Perceived importance for message quality or diversity & \cite{wangInterpretableDirectedDiversity2022} \\
Perceived stress & \cite{kimPredictionRetrospectionIntegrating2022} \\
Perceived autonomy & \cite{maWhoShouldTrust2023} \\
Persuasiveness & \cite{guesmiExplainingUserModels2022} \\
Predictability & \cite{paniguttiCodesignHumancenteredExplainable2023} \\
Rating rationale & \cite{wangRePromptAutomaticPrompt2023} \\
Recognizability & \cite{sabuncuogluDevelopingMultimodalClassroom2023} \\
Representativeness & \cite{zhangDebiasedCAMMitigateImage2022} \\
Rule consultation & \cite{raymondCultureBasedExplainableHumanAgent2020} \\
Scrutability & \cite{guesmiExplainingUserModels2022} \\
Social influence & \cite{paniguttiCodesignHumancenteredExplainable2023} \\
Usage behavior & \cite{alsulamiExploringUsersPerception2022} \\
Understanding of stress & \cite{kimPredictionRetrospectionIntegrating2022} \\
Visualization familiarity & \cite{hernandez-bocanegraExplainingRecommendationsConversations2023}, \cite{guesmiExplainingUserModels2022} \\
Willingness to pay & \cite{bendavidExplainableAIAdoption2021} \\

\end{longtable}

%% file: Tables/XAI_Properties_Concepts_Explanations.tex
\begin{longtable}{m{0.3\linewidth}m{0.3\linewidth}m{0.3\linewidth}}
\caption{XAI Properties Concepts}
\label{tab:XAI_properties_concepts}\\
\hline
\textbf{Concept} & \textbf{Explanation} & \textbf{Example} \\
\hline
Analysis/Assessment System & A system that provides analyzed and/or assessed information. & A tool that categorizes the content of drawings. \cite{caiEffectsExamplebasedExplanations2019} \\
\hline
Decision Support System & A system that helps making informed decisions by analyzing, assessing and providing relevant information. & A navigation system recommending various routes. \cite{chazetteRequirementsExplanationsQuality2022} \\
\hline
Gaming System & A system used for games and gaming. & A tool that enables users to play a sales game and provides recommendations. \cite{bendavidExplainableAIAdoption2021} \\
\hline
Recommender System & A system that provides suggestions. & Recommended movies based on user's review of other movies. \cite{donkersExplainingRecommendationsMeans2020} \\
\hline
Graphical & The explanation is provided graphically. & Colors signaling the importance of a feature for output generation. \cite{chenUnderstandingRoleHuman2023} \\
\hline
Numerical & The explanation is provided in numbers. & Confidence scores or percentages. \cite{kimAlphaDAPRAIbasedExplainable2023} \\
\hline
Textually & The explanation is provided textual (spoken and/or written). & AI-generated chat answer. \cite{youSelfdiagnosisHowChatbotbased2023} \\
\hline
Local & Explanation about specific output. & Marking words that have the highest influence on the output in an analyzed text. \cite{nielsenEffectsXAILegal2023} \\
\hline
Global & Explanation about general model behavior. & Description of algorithms behind the AI. \cite{nakaoInvolvingEndusersInteractive2022} \\
\hline
Contrastive/\newline
Counterfactual & A system's output is explained by why it occurred and not another. & Why option A rather than option B. \cite{krarupContrastiveExplanationsPlans2021} \\
\hline
Example-based & A system's output is explained by examples. & Examples from the training data that resulted in a similar or identical output. \cite{ohUnderstandingHowPeople2020} \\
\hline
Rule-based & A system's output is explained by rules. & The models' decision tree. \cite{mishraWhyWhyNot2022} \\
\hline
Causal & A system's output is explained by causalities. & Option A was chosen to cause situation B. \cite{ehsanAutomatedRationaleGeneration2019} \\
\hline
Feature-based & The output of a system is explained by revealing the features that influenced the output. & Highlighting words in biography texts that were relevant for a job description. \cite{chenUnderstandingRoleHuman2023} \\
\hline
Interactive & Interactions with the systems explanations are possible. & Users can ask to specify the explanation. \cite{chazetteRequirementsExplanationsQuality2022} \\
\hline
Static & For the user, it is not possible to interact with the systems explanations. & The explanation is informative only. \cite{guerdanAffectiveXAIFacial2021} \\
\hline
\end{longtable}

%% file: Tables/Evaluation_Concepts_Explanations.tex
\begin{longtable}{m{0.3\linewidth}m{0.3\linewidth}m{0.3\linewidth}}
\caption{XAI Evaluation Concepts}
\label{tab:evaluation_concepts}\\
\hline
\textbf{Concept} & \textbf{Explanation} & \textbf{Example} \\
\hline
Satisfaction & Users' general experience and opinion of the system. & Ratings, e.g., satisfaction with the information received from the system. \cite{kimAlphaDAPRAIbasedExplainable2023} \\
\hline
Trust & Users' reliance that the system and its output is accurate. & Ratings, e.g., faith in the provided information \cite{kimAlphaDAPRAIbasedExplainable2023}, or behavior related to trust, e.g., willingness to follow a recommendation. \cite{yangHowVisualExplanations2020} \\
\hline
Mental Model & User's conceptual understanding of how the system functions, including its purpose. & Users have to explain their understanding of the algorithms. \cite{kimPredictionRetrospectionIntegrating2022} \\
\hline
Understandability & Perceived ease to understand how the system functions. & Ratings, e.g., ability to comprehend what the system does. \cite{kimAlphaDAPRAIbasedExplainable2023} \\
\hline
Task load & Perceived cognitive load while using the system. & Ratings, e.g., cognitive effort. \cite{kimImprovedExplanatoryEfficacy2020} \\
\hline
Ease of\newline
Use/Learning & Perceived cognitive effort to use and learn how to use the system. & Ratings, e.g., overall ease to use the system. \cite{kimAlphaDAPRAIbasedExplainable2023} \\
\hline
Effectiveness & Perceived performance quality and success of the system. & Ratings, e.g., how well an AI bot plays a game. \cite{mirandaInteractiveExplainableCaseBased2021} \\
\hline
Efficiency & Perceived saved time/effort caused by the systems help. & Ratings, e.g., acquisition of quick suggestions. \cite{youSelfdiagnosisHowChatbotbased2023} \\
\hline
Usefulness & Perceived usefulness of the system. & Ratings, e.g., how much the system supports the user to accomplish a task. \cite{kimAlphaDAPRAIbasedExplainable2023} \\
\hline
Intention to use & Users willingness to use the system again. & Ratings, e.g., the intention to use the system frequently. \cite{kimAlphaDAPRAIbasedExplainable2023} \\
\hline
Transparency & Perceived clarity of the relationship between feature values and the model's decision. & Ratings, e.g., sufficiency of provided explanations. \cite{youSelfdiagnosisHowChatbotbased2023} \\
\hline
Explanation usefulness & Perceived usefulness of explanations to understand the system. & Ratings, e.g., helpfulness of highlighted words indicating feature importance in the output. \cite{drawsExplainableCrossTopicStance2023} \\
\hline
Explanation satisfaction & Satisfaction with the explanations of the system. & Ratings, e.g., about the goodness of explanations. \cite{bendavidExplainableAIAdoption2021} \\
\hline
AI Familiarity & Knowledge of and/or experience in AI tools. & Ratings, e.g., about comprehension of AI. \cite{kimAlphaDAPRAIbasedExplainable2023} \\
\hline
Cognitive Style & Measurements of the way people think or engage in tasks. & Enjoyment and engagement in cognitive activities. \cite{paniguttiCodesignHumancenteredExplainable2023} \\
\hline
Domain Expertise & Familiarity with task and/or domain that is addressed by the system. & Gaming experience in a gaming system evaluation. \cite{mirandaInteractiveExplainableCaseBased2021} \\
\hline
Prior Trust & General trust in AI or automation prior to the study. & Ratings, e.g., prior trust in machine learning. \cite{weideleAutoDOVizHumanCenteredAutomation2023} \\
\hline
User Behavior & User actions and interactions that do not directly affect task performance. User behavior can also include biometrical measures. & Users' tendency to skip or access explanations.\cite{naisehNudgingFrictionApproach2021}; Users' EEG response when interacting with the system \cite{kimImprovedExplanatoryEfficacy2020} \\
\hline
User Task Performance & User actions and interactions that are performance-related. & Average decision time in a decision-making task. \cite{jesusHowCanChoose2021} \\
\hline
\caption*{\small \textit{Note:} Proposed explanations differ from these in table \ref{tab:target_properties_xai}. They are used to facilitate the discussion of concepts in the reviewed literature.}
\end{longtable}

%% file: Tables/Concept_Matrix_Properties.tex
\centering
\preto\tabular{\setcounter{magicrownumbers}{0}}
\settowidth\rotheadsize{Analysis/Assessment System}
\begin{longtblr}[
  caption={XAI Properties Table},
  label={}
  ]{
    hline{1,Z} = 1pt,
    hline{2-Y}=solid, vlines,
    colspec = {X[10,l] *{4}{X[1,c]} *{3}{X[1,c]} *{2}{X[1,c]} *{5}{X[1,c]} *{2}{X[1,c]}}, 
    cell{3}{2-Z} = {cmd=\rotcell}, 
    row{1} = {font=\bfseries}, 
    vline{2,6, 9, 11, 16} = {2pt}, 
    rowhead=3, 
  }
			\SetCell[r=2]{f, font=\bfseries} Articles
			& \SetCell[c=4]{c} Core System 
			& & & 
			& \SetCell[c=12]{c} Explanations 
			& & & & & & & & & &
			\\
			\SetCell[r=2]{f, font=\bfseries} Articles
			& \SetCell[c=4]{c} Systems 
			& & &
			& \SetCell[c=3]{c} Modality 
			& &  
			& \SetCell[c=2]{c} Scope 
			& 
			& \SetCell[c=5]{c} Approach 
			& & & &
			 & \SetCell[c=2]{c} \makecell{Engage-\\ ment} 
			& 
			\\
            & Analysis System 
            & Decision Support System 
            & Gaming System 
			& Recommender System 
			& Graphical 
      		& Numerical 
			& Textual 
			& Local 
			& Global 
   			& Causal 
			& Contrastive/Counterfactual 
			& Example-based 
   			& Feature-based 
			& Rule-based 
			& Interactive 
			& Static 
			\\
         \rownumber~Alsulami\cite{alsulamiExploringUsersPerception2022}
			&\TB & & &
			&  & &\TB 
			&\TB & 
			& & & & & 
			&  &\TB  
			\\
            \rownumber~Arrotta\cite{arrottaDeXARDeepExplainable2022}
			&\TB & & &
			&  & &\TB 
			&\TB & 
			&\TB & & & & 
			&  &\TB  
			\\ 
            \rownumber~Ben David\cite{bendavidExplainableAIAdoption2021}
			& & &\TB &
			& &\TB &\TB 
			&\TB &\TB 
			& & & &\TB & 
			&  &\TB  
			\\ 
            \rownumber~Bhattacharya \newline
            \cite{bhattacharyaDirectiveExplanationsMonitoring2023}
			& & & &\TB 
			&\TB &\TB &\TB 
			&\TB & 
			& &\TB & &\TB &\TB 
			&\TB  &  
			\\ 
            \rownumber~Cai\cite{caiEffectsExamplebasedExplanations2019}
			&\TB & & & 
			&\TB & &\TB 
			&\TB &\TB 
			& & &\TB & & 
			& &\TB  
			\\ 
            \rownumber~Chandramouli\newline
    \cite{chandramouliInteractivePersonalizationClassifiers2023}
			&\TB & & & 
			&\TB & & 
			&\TB & 
			& & & &\TB & 
			& &\TB 
			\\ 
            \rownumber~Chazette\cite{chazetteRequirementsExplanationsQuality2022}
			& &\TB & & 
			& & &\TB 
			&\TB & 
			& & & &\TB & 
			&\TB & 
			\\ 
            \rownumber~Chen\cite{chenUnderstandingRoleHuman2023}
			& &\TB & & 
			&\TB & &\TB 
			&\TB & 
			& & &\TB &\TB & 
			& &\TB 
			\\ 
            \rownumber~Danry\cite{danryWearableReasonerEnhanced2020}
			&\TB & & & 
			& & &\TB 
			&\TB & 
			&\TB & & & & 
			& &\TB 
			\\ 
            \rownumber~Dodge\cite{dodgeExplainingModelsEmpirical2019}
			&\TB & & & 
			& &\TB &\TB 
			&\TB &\TB 
			&\TB &\TB & &\TB &\TB 
			& &\TB 
			\\ 
            \rownumber~Donkers\cite{donkersExplainingRecommendationsMeans2020}
			& & & &\TB 
			&\TB & &\TB 
			&\TB & 
			& & & &\TB & 
			& &\TB 
			\\ 
            \rownumber~Draws\cite{drawsExplainableCrossTopicStance2023}
			&\TB & & & 
			&\TB & &\TB 
			&\TB & 
			& & & &\TB & 
			& &\TB 
			\\ 
            \rownumber~Ehsan\cite{ehsanAutomatedRationaleGeneration2019}
			& & &\TB & 
			& & &\TB 
			&\TB & 
			&\TB & & & & 
			& &\TB 
			\\ 
            \rownumber~Eiband\cite{eibandBringingTransparencyDesign2018}
			& & & &\TB 
			& & &\TB 
			&\TB & 
			& & & &\TB & 
			&\TB & 
			\\ 
            \rownumber~Ghai\cite{ghaiExplainableActiveLearning2021}
			&\TB & & & 
			&\TB & & 
			&\TB & 
			& & & &\TB & 
			&\TB & 
			\\
            \rownumber~Ghassabi\cite{ghassabiLeveragingKnowledgeGraphs2023}
			&\TB & & & 
			& & &\TB 
			&\TB & 
			& & & & & 
			& &\TB 
			\\
            \rownumber~Guerdan\cite{guerdanAffectiveXAIFacial2021}
			& & &\TB & 
			& & & 
			& & 
			& & & & & 
			& &\TB 
			\\
            \rownumber~Guesmi\cite{guesmiExplainingUserModels2022}
			& & & &\TB 
			&\TB &\TB &\TB 
			&\TB &\TB 
			& & & &\TB & 
			&\TB & 
			\\\rownumber~Guesmi\cite{guesmiOndemandPersonalizedExplanation2021}
			& & & &\TB 
			&\TB &\TB &\TB 
			&\TB &\TB 
			& & & &\TB & 
			&\TB & 
			\\
            \rownumber~Guo\cite{guoEffectRecommendationSource2022}
			& & & &\TB 
			&\TB & &\TB 
			&\TB & 
			& & & &\TB & 
			& &\TB 
			\\
            \rownumber~Hanif\cite{hanifEvidenceBasedPipeline2022}
			&\TB & & & 
			&\TB &\TB & 
			&\TB & 
			& & & & & 
			&\TB & 
			\\
            \rownumber~Hernandez-Bocanegra\cite{hernandez-bocanegraExplainingRecommendationsConversations2023}
			& & & &\TB 
			& &\TB &\TB 
			&\TB & 
			&\TB & & &\TB & 
			&\TB & 
			\\
            \rownumber~Hong\cite{hongVisualizingComparingMachine2023}
			& &\TB & & 
			&\TB & &\TB 
			& &\TB 
			& & & &\TB & 
			&\TB & 
			\\
   \rownumber~Ibrahim\cite{ibrahimExplanationsImproveQuality2023}
			& &\TB & & 
			& & &\TB 
			&\TB & 
			& &\TB & &\TB & 
			& &\TB 
			\\
    \rownumber~Jesus\cite{jesusHowCanChoose2021}
			&\TB & & & 
			&\TB &\TB &\TB 
			&\TB & 
			& & & &\TB & 
			& &\TB 
			\\
       \rownumber~Kim\cite{kimImprovedExplanatoryEfficacy2020}
			& & & &\TB 
			&\TB & & 
			&\TB & 
			& & & &\TB & 
			&\TB & 
			\\
        \rownumber~Kim\cite{kimAlphaDAPRAIbasedExplainable2023}
			&\TB & & & 
			&\TB &\TB & 
			&\TB & 
			& & &\TB & & 
			& &\TB 
			\\
        \rownumber~Kim\cite{kimBubbleuExploringAugmented2023}
			& & &\TB & 
			&\TB & &\TB 
			&\TB & 
			& & & & & 
			&\TB & 
			\\
           \rownumber~Kim\cite{kimPredictionRetrospectionIntegrating2022}
			&\TB & & & 
			&\TB & &\TB 
			&\TB & 
			& & & &\TB & 
			& &\TB 
			\\
           \rownumber~Kim\cite{kimExoskeletonMindExploring2023}
			&\TB & & & 
			& & &\TB 
			&\TB & 
			&\TB & & & & 
			&\TB & 
			\\
            \rownumber~Krarup\cite{krarupContrastiveExplanationsPlans2021}
			& &\TB & & 
			&\TB & &\TB 
			&\TB & 
			&\TB &\TB & & & 
			&\TB & 
			\\
   \rownumber~Kusuma\cite{kusumaCivilWarTwin2022}
			&\TB & & & 
			&\TB & &\TB 
			&\TB & 
			& & &\TB & & 
			& &\TB 
			\\
      \rownumber~Lee\cite{leeLIMEADEAIExplanations2023}
			& & & &\TB 
			& & &\TB 
			&\TB & 
			& & & &\TB & 
			&\TB & 
			\\
         \rownumber~Liu\cite{liuIncreasingUserTrust2022}
			&\TB & & & 
			&\TB & & 
			&\TB & 
			& & & &\TB & 
			&\TB & 
			\\
            \rownumber~Ma\cite{maWhoShouldTrust2023}
			& &\TB & & 
			&\TB &\TB & \TB
			&\TB & 
			& & & &\TB & 
			& &\TB 
			\\
            \rownumber~Miranda\cite{mirandaInteractiveExplainableCaseBased2021}
			& & &\TB & 
			&\TB &\TB & 
			&\TB &\TB 
			& &\TB & & & 
			&\TB & 
			\\
            \rownumber~Mishra\cite{mishraWhyWhyNot2022}
			&\TB & & & 
			&\TB & & 
			&\TB & 
			&\TB &\TB & & & 
			&\TB & 
			\\
            \rownumber~Mishra\cite{mishraCrowdsourcingEvaluatingConceptdriven2021}
			&\TB & & & 
			&\TB & &\TB 
			&\TB &\TB 
			&\TB & & &\TB & 
			&\TB & 
			\\
            \rownumber~Naiseh\cite{naisehNudgingFrictionApproach2021}
			& &\TB & & 
			&\TB & & 
			&\TB & 
			& & & &\TB & 
			& &\TB 
			\\
            \rownumber~Nakao\cite{nakaoInvolvingEndusersInteractive2022}
			&\TB & & & 
			&\TB &\TB &\TB 
			&\TB &\TB 
			& & &\TB &\TB & 
			&\TB & 
			\\
            \rownumber~Nielsen\cite{nielsenEffectsXAILegal2023}
			&\TB & & & 
			&\TB & & 
			&\TB & 
			& & & &\TB & 
			& &\TB 
			\\
            \rownumber~Oh\cite{ohUnderstandingHowPeople2020}
			&\TB  & & & 
			&\TB & &\TB 
			&\TB  & 
			& & &\TB  & & 
			& &\TB  
			\\
            \rownumber~Panigutti\cite{paniguttiCodesignHumancenteredExplainable2023}
			& &\TB & & 
			&\TB & &\TB 
			&\TB & 
			& & & & &\TB 
			&\TB &  
			\\
            \rownumber~Piorkowski\cite{piorkowskiAIMEEExploratoryStudy2023}
			&\TB & & & 
			&\TB &\TB &\TB 
			&\TB & 
			& & & & &\TB 
			&\TB &  
			\\
               \rownumber~Raymond\cite{raymondCultureBasedExplainableHumanAgent2020}
			& & &\TB & 
			& & &\TB 
			&\TB & 
			& &\TB & & & 
			& &\TB  
			\\
            \rownumber~Sabuncuoglu \newline\cite{sabuncuogluDevelopingMultimodalClassroom2023}
			&\TB & & & 
			& & &\TB 
			&\TB & 
			& & & & &\TB 
			& &\TB  
			\\
            \rownumber~Sajja\cite{sajjaExplainableAIBased2021}
			& &\TB & & 
			&\TB & & 
			&\TB & 
			& &\TB & & & 
			&\TB &  
			\\
            \rownumber~Sevastjanova\newline
        \cite{sevastjanovaQuestionCombGamificationApproach2021}
			&\TB & & & 
			& & &\TB 
			&\TB & 
			& & & &\TB & 
			&\TB &  
			\\
            \rownumber~Spinner\cite{spinnerExplAInerVisualAnalytics2019}
			&\TB & & & 
			&\TB & &\TB 
			&\TB &\TB 
			& & & &\TB & 
			&\TB &  
			\\
            \rownumber~Sun\cite{sunDesigningDirectFeedback2023}
			&\TB & & & 
			&\TB &\TB & 
			&\TB & 
			& & &\TB &\TB & 
			&\TB &  
			\\
            \rownumber~Tandon\cite{tandonSurfacingAIExplainability2023}
			&\TB & & & 
			&\TB &\TB &\TB 
			&\TB & 
			& & & &\TB & 
			& &\TB  
			\\
            \rownumber~Tsai\cite{tsaiExploringPromotingDiagnostic2021}
			&\TB & & & 
			& & &\TB 
			&\TB & 
			&\TB & &\TB & &\TB 
			&\TB &  
			\\
            \rownumber~Tutul\cite{tutulInvestigatingTrustHumanMachine2021}
			&\TB & & & 
			&\TB & & 
			&\TB &\TB 
			& & & &\TB & 
			&\TB &  
			\\
            \rownumber~Ueno\cite{uenoTrustRelianceConsensusBased2023}
			&\TB & & & 
			& & &\TB 
			&\TB & 
			& & & & & 
			&\TB &  
			\\
            \rownumber~Wang\cite{wangRePromptAutomaticPrompt2023}
			&\TB & & & 
			& & &\TB 
			&\TB & 
			& & & & & 
			& &\TB  
			\\
            \rownumber~Wang\cite{wangInterpretableDirectedDiversity2022}
			&\TB & & & 
			&\TB &\TB &\TB 
			&\TB & 
			& &\TB & &\TB & 
			&\TB &  
			\\
   \rownumber~Weidele\cite{weideleAutoDOVizHumanCenteredAutomation2023}
			&\TB & & & 
			&\TB &\TB &\TB 
			&\TB & 
			& & & & &\TB 
			&\TB &  
			\\

   \rownumber~Xu\cite{xuXAIRFrameworkExplainable2023}
			& & & &\TB 
			&\TB & &\TB 
			&\TB & 
			&\TB & & & & 
			&\TB &  
			\\

      \rownumber~Yang\cite{yangHowVisualExplanations2020}
			&\TB & & & 
			&\TB &\TB & 
			&\TB & 
			& & &\TB & & 
			& &\TB  
			\\
         \rownumber~Ye\cite{yeWikipediaORESExplorer2021}
			&\TB & & & 
			&\TB &\TB &\TB 
			&\TB &\TB 
			& & & & & 
			&\TB &  
			\\
            \rownumber~You\cite{youSelfdiagnosisHowChatbotbased2023}
			& &\TB & & 
			& & &\TB 
			&\TB & 
			&\TB & & & & 
			&\TB &  
			\\
             \rownumber~Yu\cite{yuKeepingDesignersLoop2020}
			&\TB & & & 
			&\TB & &\TB 
			&\TB & 
			& & & & & 
			&\TB &  
			\\
            \rownumber~Zhang\cite{zhangAdaVisAdaptiveExplainable2023}
			& & & &\TB 
			& & &\TB 
			& &\TB 
			& & & & &\TB 
			& &\TB  
            \\
            \rownumber~Zhang\cite{zhangDebiasedCAMMitigateImage2022}
			&\TB & & & 
			&\TB & &\TB 
			&\TB & 
			& & & &\TB & 
			& &\TB  
			\\
            \rownumber~Zöller\cite{zollerXAutoMLVisualAnalytics2023}
			&\TB & & & 
			&\TB &\TB & 
			&\TB & 
			& & & & & 
			&\TB &  
			\\
            $\sum$
			&38 &10 &6 &11 
			&43 &20 &49  
			&42 &13 
			&12 &9 &9 &33 &8 
			&35 &30 
			\\
   		\end{longtblr}

%% file: Tables/Concept_Matrix_Evaluation.tex
\centering
\preto\tabular{\setcounter{magicrownumbers}{0}}
\settowidth\rotheadsize{Contrastive/Counterfactual}
\newgeometry{left=1cm,right=1cm,top=1.5cm,bottom=1.5cm}
\begin{landscape}
\begin{longtblr}[
  caption = {Metrics Table},
  label = {},
]{  
  width = \textwidth,
  colspec = {X[13,l] *{9}{X[2,c]} *{2}{X[2,c]} *{2}{X[1,c]} *{4}{X[2,c]} *{2}{X[1,c]}},
  hline{1,Z} = 1pt,
  hline{2-Y} = solid,
  vlines,
  cell{4}{2-Z} = {cmd=\rotcell},
  row{1} = {font=\bfseries},
  row{3} = {font=\itshape},
  vline{2,4,7,12,13,15,19} = {2pt},
  rowhead = 4,
}
		
			\SetCell[r=2]{f, font=\bfseries} \parbox{2cm}{Number/\\Article}
			
			& \SetCell[c=13]{c} System Metrics 
			& & & & & & & & & & & & 
			& \SetCell[c=6]{c} User Metrics 
			& & & & &
			\\
			\SetCell[r=2]{f, font=\bfseries} Articles
			& \SetCell[c=11]{c} Core System 
			& & & & & & & & & &
            & \SetCell[r=2, c=2]{c} \makecell{Expla-\\ nation} 
			& 
			& \SetCell[r=2, c=4]{c} Characteristics 
			& & &
			& \SetCell[r= 2, c=2]{c} \makecell{Acti-\\ vity} 
            &
			\\
            & \SetCell[c=2]{c} \makecell{Affect-\\ ion}
            &
            & \SetCell[c=3]{c} Cognition
            & &
            & \SetCell[c=5]{c} Usability
            & & & &
            & \SetCell[c=1]{c} \makecell{Inter- \\ preta- \\ bility}
            \\

         	& Satisfaction 
			& Trust 
         	& Mental Model 
   			& Understandability 
      		& Task load 
            & Ease of Use and Learning 
   			& Effectiveness 
            & Efficiency 
      		& Usefulness 
            & Intention to use 
            & Transparency 
   			& Explanation usefulness 
			& Explanation satisfaction 
			& AI Familiarity 
            & Cognitive Style 
            & Domain Expertise 
			& Prior Trust 
            & User Behavior
			& User Task Performance 
			\\
\rownumber~Alsulami\cite{alsulamiExploringUsersPerception2022}
			&  & &  & &  & &  & & &  & 
			& & 
			& & &\TB  &
			&\TB   & \\ 
            \rownumber~Arrotta\cite{arrottaDeXARDeepExplainable2022}
			&  & &  & &  & &  & & &  & 
			& &\TB 
			& & &  &
			&  & \\ 
            \rownumber~Ben David\cite{bendavidExplainableAIAdoption2021}
			& &\TB & & & & & & & &\TB & 
			& &\TB 
			& & &  &
			&  & \\ 
            \rownumber~Bhattacharya\cite{bhattacharyaDirectiveExplanationsMonitoring2023}
			& &\TB &\TB &\TB & & & & &\TB &\TB  & 
			& & 
			& & &  &
			&\TB  &\TB \\ 
            \rownumber~Cai\cite{caiEffectsExamplebasedExplanations2019}
			& &\TB & &\TB & & & & & & & 
			& & 
			& & & &
			&\TB & \\ 
            \rownumber~Chandramouli\cite{chandramouliInteractivePersonalizationClassifiers2023}
			& & & & & & & & & & & 
			& & 
			& & & &
			& & \\ 
            \rownumber~Chazette\cite{chazetteRequirementsExplanationsQuality2022}
			&\TB & & & & & & & & & & 
			&\TB &\TB 
			& & &\TB &
			&\TB & \\ 
            \rownumber~Chen\cite{chenUnderstandingRoleHuman2023}
			& & & & & & & & & &\TB & 
			& & 
			&\TB & & &
			&\TB &\TB \\ 
            \rownumber~Danry\cite{danryWearableReasonerEnhanced2020}
			& & & & & &\TB & & &\TB & & 
			& & 
			& & & &
			& & \\ 
   
\rownumber~Dodge\cite{dodgeExplainingModelsEmpirical2019}
			& & & & & & & & & & & 
			& & 
			& &\TB & &\TB 
			& & \\ 
   \rownumber~Donkers\cite{donkersExplainingRecommendationsMeans2020}
			& & & & & & & & & & & 
			&\TB &\TB 
			& & & & 
			& & \\ 
      \rownumber~Draws\cite{drawsExplainableCrossTopicStance2023}
			& & &\TB & & & & & & & & 
			&\TB &\TB 
			& & & & 
			& &\TB \\ 
         \rownumber~Ehsan\cite{ehsanAutomatedRationaleGeneration2019}
			& &\TB & &\TB & & & & & & & 
			& &\TB 
			& & & & 
			& & \\ 
            \rownumber~Eiband\cite{eibandBringingTransparencyDesign2018}
			& & &\TB & & &\TB & & & &\TB & 
			& & 
			& & & & 
			& & \\ 
            \rownumber~Ghai\cite{ghaiExplainableActiveLearning2021}
			&\TB &\TB & & &\TB & & & & & & 
			& & 
			&\TB &\TB &\TB & 
			& &\TB \\ 
            \rownumber~Ghassabi\cite{ghassabiLeveragingKnowledgeGraphs2023}
			& &\TB & &\TB & & & & & & & 
			&\TB &\TB 
			&\TB & &\TB & 
			& & \\ 
            \rownumber~Guerdan\cite{guerdanAffectiveXAIFacial2021}
			& & & & & & & & & & & 
			& & 
			& & & & 
			&\TB &\TB \\ 
            \rownumber~Guesmi\cite{guesmiExplainingUserModels2022}
			& &\TB & & & & &\TB &\TB & & &\TB 
			& &\TB 
			&\TB &\TB &\TB &\TB 
			& & \\ 
\rownumber~Guesmi\cite{guesmiOndemandPersonalizedExplanation2021}
			& & & & & & & & & & & 
			& &\TB 
			&\TB & &\TB & 
			& & \\ 
            \rownumber~Guo\cite{guoEffectRecommendationSource2022}
			&\TB &\TB & &\TB & & &\TB & &\TB & &\TB 
			& & 
			& & & & 
			& & \\ 
            \rownumber~Hanif\cite{hanifEvidenceBasedPipeline2022}
			& &\TB & &\TB & & & & &\TB & & 
			& & 
			& & &\TB & 
			& & \\ 
            \rownumber~Hernandez-Bocanegra\cite{hernandez-bocanegraExplainingRecommendationsConversations2023}
			&\TB &\TB &\TB &\TB & & &\TB & &\TB &\TB &\TB 
			&\TB &\TB 
			& &\TB & & 
			& & \\ 
            \rownumber~Hong\cite{hongVisualizingComparingMachine2023}
			& &\TB & & & &\TB & & & &\TB & 
			& & 
			&\TB & &\TB & 
			&\TB &\TB \\ 
   \rownumber~Ibrahim\cite{ibrahimExplanationsImproveQuality2023}
			& &\TB & & & & & & & & & 
			&\TB & 
			&\TB & &\TB & 
			& &\TB\\ 
      \rownumber~Jesus\cite{jesusHowCanChoose2021}
			& & & & & & & & & & & 
			&\TB & 
			& & & & 
			& &\TB\\ 
        \rownumber~Kim\cite{kimImprovedExplanatoryEfficacy2020}
			& & & & &\TB & & & & & & 
			&\TB & 
			& & & & 
			&\TB &\\ 
        \rownumber~Kim\cite{kimAlphaDAPRAIbasedExplainable2023}
			&\TB &\TB &\TB &\TB & &\TB & & &\TB &\TB & 
			& & 
			&\TB & & & 
			& &\\ 
       \rownumber~Kim\cite{kimBubbleuExploringAugmented2023}
			& & & & &\TB & & & & & & 
			& & 
			& & &\TB & 
			& &\\ 
          \rownumber~Kim\cite{kimPredictionRetrospectionIntegrating2022}
			& &\TB &\TB & &\TB &\TB &\TB &\TB &\TB & & 
			&\TB & 
			& & &\TB & 
			&\TB &\\ 
          \rownumber~Kim\cite{kimExoskeletonMindExploring2023}
			& & & & & &\TB & & &\TB &\TB & 
			& & 
			& & &\TB & 
			&\TB &\\ 
     \rownumber~Krarup\cite{krarupContrastiveExplanationsPlans2021}
			& & & & & & & & & & & 
			&\TB &\TB 
			& & & & 
			& &\\ 
   \rownumber~Kusuma\cite{kusumaCivilWarTwin2022}
			& & & & & & & & & & & 
			& & 
			&\TB & &\TB & 
			& &\\ 
      \rownumber~Lee\cite{leeLIMEADEAIExplanations2023}
			&\TB &\TB & & & &\TB &\TB & & &\TB &\TB 
			& & 
			& & &\TB & 
			& &\\ 
         \rownumber~Liu\cite{liuIncreasingUserTrust2022}
			&\TB &\TB & &\TB & & &\TB & &\TB & & 
			& & 
			& & &\TB & 
			&\TB &\\ 
            \rownumber~Ma\cite{maWhoShouldTrust2023}
			&\TB &\TB & & &\TB &\TB & & &\TB &\TB & 
			&\TB & 
			&\TB & & & 
			& &\TB\\ 
            \rownumber~Miranda\cite{mirandaInteractiveExplainableCaseBased2021}
			& & & &\TB & & &\TB & &\TB &\TB & 
			&\TB & 
			&\TB & &\TB & 
			& &\\ 
            \rownumber~Mishra\cite{mishraWhyWhyNot2022}
			&\TB &\TB & &\TB & &\TB & & &\TB & & 
			& &\TB 
			&\TB & &\TB & 
			& &\\ 
            \rownumber~Mishra\cite{mishraCrowdsourcingEvaluatingConceptdriven2021}
			& &\TB & & & & &\TB & & & & 
			& &\TB 
			& & & & 
			& &\\ 
            \rownumber~Naiseh\cite{naisehNudgingFrictionApproach2021}
			& &\TB & & & & & & & & & 
			& & 
			& & &\TB & 
			&\TB &\\ 
            \rownumber~Nakao\cite{nakaoInvolvingEndusersInteractive2022}
			& & & & &\TB & & & & & & 
			& & 
			&\TB & & &\TB 
			& &\\ 
            \rownumber~Nielsen\cite{nielsenEffectsXAILegal2023}
			& & & & & & & & & & & 
			& & 
			& & & & 
			&\TB &\\ 
            \rownumber~Oh\cite{ohUnderstandingHowPeople2020}
			& & & &\TB & & & & & & & 
			& & 
			&\TB & &\TB & 
			& &\\ 
            \rownumber~Panigutti\cite{paniguttiCodesignHumancenteredExplainable2023}
			& &\TB & &\TB & &\TB &\TB &\TB &\TB & & 
			& &\TB 
			& &\TB &\TB &\TB 
			& &\\ 
            \rownumber~Piorkowski\cite{piorkowskiAIMEEExploratoryStudy2023}
			&\TB & &\TB &\TB & &\TB & & &\TB & & 
			& & 
			&\TB & &\TB & 
			& &\\ 
            \rownumber~Raymond\cite{raymondCultureBasedExplainableHumanAgent2020}
			& & & & & & & & & & & 
			& & 
			& & & & 
			& &\TB\\ 
            \rownumber~Sabuncuoglu\cite{sabuncuogluDevelopingMultimodalClassroom2023}
			&\TB & & & & &\TB &\TB &\TB & & & 
			& & 
			& & & & 
			& &\\ 
            \rownumber~Sajja\cite{sajjaExplainableAIBased2021}
			& & & & & & & & & & & 
			&\TB &\TB 
			& & & & 
			& &\\ 
            \rownumber~Sevasjanova\cite{sevastjanovaQuestionCombGamificationApproach2021}
			& & & & & & & & & & & 
			& & 
			& & &\TB & 
			& &\\ 
            \rownumber~Spinner\cite{spinnerExplAInerVisualAnalytics2019}
			& & & & & & & & &\TB & & 
			& & 
			&\TB & &\TB & 
			& &\\ 
            \rownumber~Sun\cite{sunDesigningDirectFeedback2023}
			& & & & & &\TB &\TB & &\TB & & 
			&\TB & 
			&\TB & &\TB & 
			& &\\ 
            \rownumber~Tandon\cite{tandonSurfacingAIExplainability2023}
			& &\TB & &\TB & & & & & & & 
			& & 
			&\TB & &\TB & 
			& &\\ 
            \rownumber~Tsai\cite{tsaiExploringPromotingDiagnostic2021}
			&\TB &\TB &\TB & &\TB &\TB &\TB & &\TB & &\TB 
			& & 
			&\TB & & & 
			& &\\ 
            \rownumber~Tutul\cite{tutulInvestigatingTrustHumanMachine2021}
			& &\TB & & & & & & &\TB & & 
			& & 
			& & &\TB & 
			& &\\ 
            \rownumber~Ueno\cite{uenoTrustRelianceConsensusBased2023}
			& &\TB & & & & & & & & & 
			& & 
			& & &\TB & 
			&\TB &\\ 
            \rownumber~Wang\cite{wangRePromptAutomaticPrompt2023}
			& & & & & & &\TB & & & & 
			& & 
			& & &\TB & 
			& &\\ 
            \rownumber~Wang\cite{wangInterpretableDirectedDiversity2022}
			&\TB & & & & &\TB & & &\TB & & 
			&\TB & 
			& & & & 
			& &\TB\\ 
   \rownumber~Weidele\cite{weideleAutoDOVizHumanCenteredAutomation2023}
			& &\TB & & & &\TB & & &\TB & & 
			& & 
			&\TB & &\TB &\TB 
			& &\\ 
   
\rownumber~Xu\cite{xuXAIRFrameworkExplainable2023}
			& &\TB & &\TB & &\TB & & & &\TB &\TB 
			& & 
			& & & & 
			& &\\ 
      \rownumber~Yang\cite{yangHowVisualExplanations2020}
			&\TB &\TB & &\TB & & & & & & & 
			&\TB & 
			&\TB & &\TB &\TB 
			& &\TB\\ 
         \rownumber~Ye\cite{yeWikipediaORESExplorer2021}
			& & & & & & & & & & & 
			& & 
			&\TB & &\TB & 
			& &\\ 
         \rownumber~You\cite{youSelfdiagnosisHowChatbotbased2023}
			&\TB &\TB & & &\TB & &\TB &\TB & & &\TB 
			& & 
			&\TB & &\TB & 
			& &\\ 
         \rownumber~Yu\cite{yuKeepingDesignersLoop2020}
			& &\TB & &\TB  & & & & & & & 
			& & 
			&\TB & & & 
			& &\TB\\ 
            \rownumber~Zhang\cite{zhangAdaVisAdaptiveExplainable2023}
			& & & &  & & &\TB & & & & 
			&\TB & 
			& & &\TB & 
			& &\\ 
            \rownumber~Zhang\cite{zhangDebiasedCAMMitigateImage2022}
			& & & &  & & & & & & & 
			&\TB & 
			&\TB & &\TB & 
			& &\\ 
            \rownumber~Zöller\cite{zollerXAutoMLVisualAnalytics2023}
			& &\TB & &\TB & &\TB & & & &\TB & 
			& & 
			&\TB & &\TB & 
			& &\\ 
              $\sum$
			&15 &31 &8 &19 &8 &18 &15 &5 &20 &13 &7 
			&18 &15 
			&27 &5 &35 &6
			&14 &13\\ 
        \end{longtblr}
    \end{landscape}

%% file: Tables/System_Properties_Condensed.tex
\begin{longtable}{@{}llr@{}}
\caption{Summary of XAI System Properties and Evaluation Metrics (Frequency Across Reviewed Papers)}
\label{tab:summary} \\
\toprule
\textbf{Category / Dimension} & \textbf{Concept} & \textbf{No. of Papers} \\
\midrule
\endfirsthead

\multicolumn{3}{@{}l}{\textit{(continued)} \tablename\ \ref{tab:summary}} \\
\toprule
\textbf{Category / Dimension} & \textbf{Concept} & \textbf{No. of Papers} \\
\midrule
\endhead

\midrule
\multicolumn{3}{r}{\textit{Continued on next page}} \\
\endfoot

\bottomrule
\endlastfoot

\multicolumn{3}{@{}l}{\textbf{XAI System Properties}} \\
\midrule
\multicolumn{3}{@{}l}{\textit{Core System}} \\
\quad Systems & Analysis System & 38 \\
              & Decision Support System & 10 \\
              & Gaming System & 6 \\
              & Recommender System & 11 \\
\addlinespace
\multicolumn{3}{@{}l}{\textit{Explanations}} \\
\quad Modality & Graphical & 43 \\
               & Numerical & 20 \\
               & Textual & 49 \\
\addlinespace
\quad Scope & Local & 42 \\
            & Global & 13 \\
\addlinespace
\quad Approach & Causal & 12 \\
               & Contrastive/Counterfactual & 9 \\
               & Example-based & 9 \\
               & Feature-based & 33 \\
               & Rule-based & 8 \\
\addlinespace
\quad Engagement & Interactive & 35 \\
                 & Static & 30 \\
\midrule
\multicolumn{3}{@{}l}{\textbf{XAI Evaluation Metrics}} \\
\midrule
\multicolumn{3}{@{}l}{\textit{Core System}} \\
\quad Affection & Satisfaction & 15 \\
                & Trust & 31 \\
\addlinespace
\quad Cognition & Mental Model & 8 \\
                & Understandability & 19 \\
                & Task load & 8 \\
\addlinespace
\quad Usability & Ease of Use and Learning & 18 \\
               & Effectiveness & 15 \\
               & Efficiency & 5 \\
               & Usefulness & 20 \\
               & Intention to use & 13 \\
\addlinespace
\quad Interpretability & Transparency & 7 \\
\addlinespace
\multicolumn{3}{@{}l}{\textit{Explanation}} \\
\quad Explanation Quality & Explanation usefulness & 18 \\
                          & Explanation satisfaction & 15 \\
\addlinespace
\multicolumn{3}{@{}l}{\textit{User}} \\
\quad User Characteristics & AI Familiarity & 27 \\
                           & Cognitive Style & 5 \\
                           & Domain Expertise & 35 \\
                           & Prior Trust & 6 \\
\addlinespace
\quad User Behavior & User Behavior & 14 \\
                    & User Task Performance & 13 \\
\end{longtable}

\begin{minipage}{0.85\linewidth}
\footnotesize \textit{Note:} Counts are not mutually exclusive; a single system may implement multiple explanation types, and a single study may measure multiple metrics. Based on $N = 65$ reviewed papers.
\end{minipage}

%% file: Tables/Design_Goal_Concepts.tex
\begin{table}[htbp]
\centering
\caption{Design Goal Concepts}
\label{tab:designgoals-concepts}
\begin{tabular}{@{}p{2.6cm}p{2.6cm}p{4.3cm}p{4.3cm}@{}}
\toprule
\textbf{Design Goal} & \textbf{Keywords} & \textbf{Explanation} & \textbf{Example} \\
\midrule
Responsibility &
Accountability, appropriate trust, ethical, fairness, reliable use, responsibility, responsible AI, trust calibration &
Designing for appropriate or calibrated trust so that neither undertrust nor overtrust arises, incorporating ethical aspects such as fairness. &
Preventing harmful overreliance by transparently disclosing the system's inner workings and decision-making. \cite{chenUnderstandingRoleHuman2023} \\
\addlinespace
Acceptance &
Acceptance, adoption, willingness to use &
Increasing usage intention and system adoption, closely related to intention to use. &
Explanation type affecting readiness to adopt a financial algorithmic advisor. \cite{bendavidExplainableAIAdoption2021} \\
\addlinespace
User Experience &
User experience, usability, user satisfaction &
Improving the quality of the user's interaction with the system, including flow and ease of use. &
Image-based explanations yielding a better user experience than charts. \cite{yangHowVisualExplanations2020} \\
\addlinespace
Human-AI Collaboration &
Feedback (loop), (human-AI) collaboration, human-AI teaming &
Enabling interactive feedback loops between humans and the AI, and moderating collaboration between humans in different roles. &
An explainable dashboard letting data scientists tune the model via interactive features. \cite{hanifEvidenceBasedPipeline2022} \\
\addlinespace
System \& User Task Performance &
Accuracy, performance &
Enhancing the measurable performance of the system and of the user's task, via computer-centric metrics and/or user feedback. &
Ranking correct visualization choices to evaluate an adaptive recommendation system. \cite{zhangAdaVisAdaptiveExplainable2023} \\
\bottomrule
\end{tabular}
\vspace{0.5em}
\begin{minipage}{\textwidth}
\footnotesize \textit{Note:} Design goals were assigned to a user group based on their relative frequency in that group. Goals are not mutually exclusive; a single paper could address multiple goals.
\end{minipage}
\end{table}

%% file: Tables/Design_Goals_User_Groups_Frequencies.tex
\newgeometry{left=1cm,right=1cm,top=1.5cm,bottom=1.5cm}
\begin{landscape}
\renewcommand{\arraystretch}{1.3}
\begin{longtable}{
>{\raggedright\arraybackslash}m{0.28\linewidth}
>{\centering\arraybackslash}m{0.10\linewidth}
>{\centering\arraybackslash}m{0.12\linewidth}
>{\centering\arraybackslash}m{0.10\linewidth}
>{\centering\arraybackslash}m{0.12\linewidth}
>{\raggedright\arraybackslash}m{0.16\linewidth}
}
\caption{Design Goal Distribution and Group Assignment}
\label{tab:goal_distribution} \\
\toprule
 & \multicolumn{2}{c}{\textbf{Data Experts} ($n=21$)} & \multicolumn{2}{c}{\textbf{AI Novices} ($n=109$)} & \\
\cmidrule(lr){2-3}\cmidrule(lr){4-5}
\textbf{Design Goal} & Abs. & Rel. & Abs. & Rel. & \textbf{Assigned to} \\
\midrule
\endfirsthead

\multicolumn{6}{@{}l}{\textit{(continued)} \tablename\ \ref{tab:goal_distribution}} \\
\toprule
 & \multicolumn{2}{c}{\textbf{Data Experts}} & \multicolumn{2}{c}{\textbf{AI Novices}} & \\
\cmidrule(lr){2-3}\cmidrule(lr){4-5}
\textbf{Design Goal} & Abs. & Rel. & Abs. & Rel. & \textbf{Assigned to} \\
\midrule
\endhead

\midrule
\multicolumn{6}{r}{\textit{Continued on next page}} \\
\endfoot

\bottomrule
\endlastfoot

Responsible Use               & 3 & 14.29\%          & 32 & \textbf{29.36\%} & AI Novices \\
Human-AI Collaboration        & 7 & \textbf{33.33\%} & 22 & 20.18\%          & Data Experts \\
System \& User Task Perf.     & 4 & \textbf{19.05\%} & 11 & 10.09\%          & Data Experts \\
Acceptance                    & 1 & 4.76\%           & 7  & \textbf{6.42\%}  & AI Novices \\
User Experience               & 6 & 28.57\%          & 37 & \textbf{33.94\%} & AI Novices \\
\midrule
\textbf{Total appearances}    & \textbf{21} & 100\%   & \textbf{109} & 100\% & \\
\end{longtable}

\begin{minipage}{0.95\linewidth}
\footnotesize \textit{Note:} Absolute frequency is the number of papers in which a design goal appeared for the respective user group. Relative frequency is that goal's share of all goal appearances within the group. Each goal was assigned to the group with the higher relative frequency (shown in \textbf{bold}). Counts are not mutually exclusive, as a single system may pursue multiple design goals.
\end{minipage}
\end{landscape}
\restoregeometry